\documentclass[lettersize,journal]{IEEEtran}
\usepackage{amsmath,amsfonts}
\usepackage{algorithmic}
\usepackage{algorithm}
\usepackage{array}
\usepackage[caption=false,font=normalsize,labelfont=sf,textfont=sf]{subfig}
\usepackage{textcomp}
\usepackage{stfloats}
\usepackage{url}
\usepackage{verbatim}
\usepackage{graphicx}
\usepackage{cite}
\usepackage{xspace}
\newcommand{\etal}{\textit{et al.}\xspace}

\usepackage{multirow}
\usepackage{makecell}

\usepackage{bm}
\usepackage{xcolor}
\usepackage{cleveref}
\crefname{figure}{Fig.}{Figs.}
\crefname{table}{Tab.}{Tabs.}
\crefname{section}{Sec.}{Secs.}
\crefname{equation}{Eq.}{Eqs.}

\newcommand{\bolddoubleline}{
    \noalign{\hrule height 0.7pt \vspace{2pt} \hrule height 0.7pt}
}

\begin{document}

\title{ISTASTrack: Bridging ANN and SNN via ISTA Adapter for RGB-Event Tracking}

\author{Siying Liu, Zikai Wang, Hanle Zheng, Yifan Hu, Xilin Wang, Qingkai Yang, Jibin Wu,~\IEEEmembership{Member,~IEEE}, \\Hao Guo, Lei Deng,~\IEEEmembership{Senior Member,~IEEE,}
        % <-this % stops a space
\thanks{Manuscript received September XX, 2025; revised XXX, 2025.}
\thanks{This work was supported in part by the National Natural Science Foundation of China (No. 62506199, 62276151 and 62411560155), China Postdoctoral Science Foundation (No. 2025M781488), Tsinghua University Initiative Scientific Research Program, and Chinese Institute for Brain Research, Beijing. (Corresponding authors: Hao Guo and Lei Deng)}% <-this % stops a space
\thanks{Siying Liu, Hanle Zheng, Yifan Hu, and Lei Deng are with the Center for Brain Inspired Computing Research (CBICR), Department of Precision Instrument, Tsinghua University, Beijing, China (e-mail: siying-liu@mail.tsinghua.edu.cn; zhl22@mails.tsinghua.edu.cn; huyf19@mails.tsinghua.edu.cn; leideng@mail.tsinghua.edu.cn).}
\thanks{Zikai Wang and Hao Guo are with the College of Computer Science and Technology, Taiyuan University of Technology, Shanxi, China (e-mail: zikaiwang@link.tyut.edu.cn; guohao@tyut.edu.cn).}
\thanks{Xilin Wang is with the Engineering Laboratory of Power Equipment Reliability in Complicated Coastal Environments, Tsinghua Shenzhen International Graduate School, Tsinghua University, Shenzhen, China (e-mail: wang.xilin@sz.tsinghua.edu.cn).}
\thanks{Qingkai Yang is with the School of Automation, Beijing Institute of Technology, Beijing, China (e-mail: qingkai.yang@bit.edu.cn).}
\thanks{Jibin Wu is with the Department of Data Science and Artificial Intelligence and the Department of Computing, The Hong Kong Polytechnic University, Hong Kong SAR (e-mail: jibin.wu@polyu.edu.hk).}
}

\markboth{Journal of \LaTeX\ Class Files,~Vol.~xx, No.~xx, September~2025}%
{Shell \MakeLowercase{\textit{et al.}}: A Sample Article Using IEEEtran.cls for IEEE Journals}


\maketitle

\begin{abstract}
RGB-Event tracking has become a promising trend in visual object tracking to leverage the complementary strengths of both RGB images and dynamic spike events for improved performance. However, existing artificial neural networks (ANNs) struggle to fully exploit the sparse and asynchronous nature of event streams. Recent efforts toward hybrid architectures combining ANNs and spiking neural networks (SNNs) have emerged as a promising solution in RGB-Event perception, yet effectively fusing features across heterogeneous paradigms remains a challenge. In this work, we propose ISTASTrack, the first transformer-based \textbf{A}NN-\textbf{S}NN hybrid \textbf{Track}er equipped with \textbf{ISTA} adapters for RGB-Event tracking. The two-branch model employs a vision transformer to extract spatial context from RGB inputs and a spiking transformer to capture spatio-temporal dynamics from event streams. To bridge the modality and paradigm gap between ANN and SNN features, we systematically design an ISTA adapter for bidirectional feature interaction between the two branches. The ISTA adapter is derived from the sparse representation theory by unfolding the iterative shrinkage-thresholding algorithm. Additionally, we incorporate a temporal downsampling attention module within the adapter to align multi-step SNN features with single-step ANN features in the latent space. Experimental results on RGB-Event tracking benchmarks, such as FE240hz, VisEvent, COESOT, and FELT, have demonstrated that ISTASTrack achieves state-of-the-art performance while maintaining high energy efficiency. This work highlights the effectiveness and practicality of hybrid ANN-SNN designs for robust visual tracking. The code is publicly available at \url{https://github.com/lsying009/ISTASTrack.git}.
\end{abstract}


\section{Introduction}
\label{sec:introduction}
% \IEEEPARstart{T}{his} file is 

Event cameras introduce a novel paradigm for visual perception by asynchronously detecting pixel-level intensity changes and generating dynamic spike events \cite{gallego_eventbased_2020}. Their unique sensing mechanism offers advantages such as high temporal resolution, high dynamic range, and low power consumption, making them well suited for fast motion and extreme lighting conditions. However, event data lack the rich spatial details that traditional RGB cameras provide, such as texture and color. Therefore, RGB-Event fusion has shown great potential in advancing a wide range of computer vision tasks, such as object detection \cite{gehrig_dsec_2021}, visual tracking \cite{liu_combined_2016}, and action recognition \cite{wang_hardvs_2024}. In this study, we focus specifically on visual object tracking (VOT), where effectively exploiting the complementary strengths of RGB and event modalities remains a key challenge.

\begin{figure}[tb]
\centering 
{\includegraphics[width=0.9\linewidth]{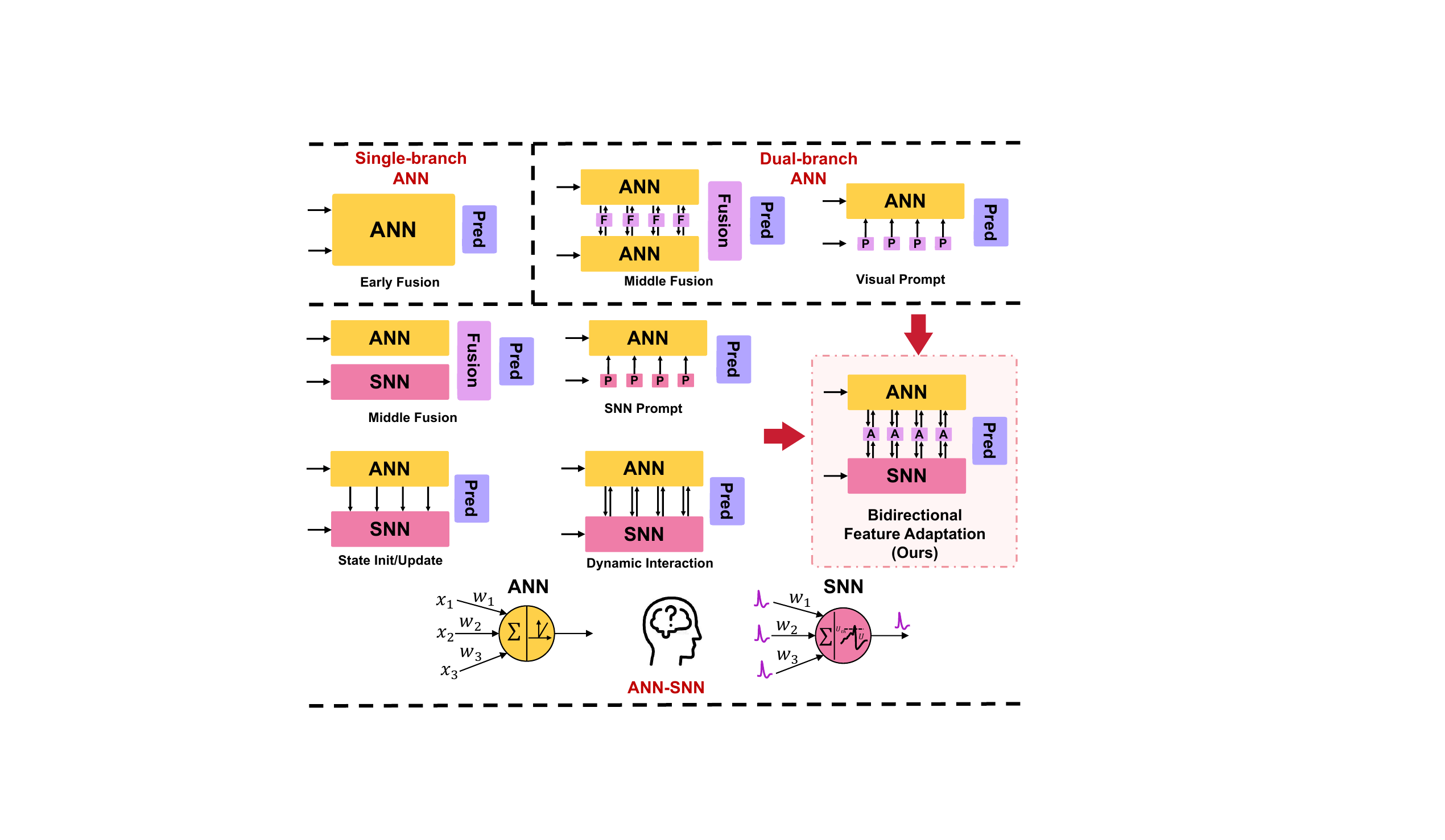}}
\caption{ANN and hybrid ANN-SNN architecture with different fusion strategies in RGB-Event perception. Our ISTASTrack exploits bidirectional feature interaction between the ANN and SNN branches.}
\label{fig:intro-dual-branch}        
\end{figure}

Most existing RGB-Event tracking frameworks employ artificial neural networks (ANNs) to process both RGB and event modalities. As illustrated in \cref{fig:intro-dual-branch}, they can be broadly divided into single-branch and dual-branch architectures. Classic single-modality trackers such as DiMP \cite{bhat_learning_2019} and OSTrack \cite{ye_joint_2022} can be extended to multi-modal settings through early fusion. CEUTrack \cite{tang_revisiting_2026} further adapts the vision transformer (ViT) framework by embedding RGB-Event inputs into a shared token space. On the other hand, dual-branch models employ either symmetric ANN backbones or modality-specific backbones, followed by dedicated middle fusion modules. Recent RGB-Event tracking approaches, including FENet \cite{zhang_object_2021}, AFNet \cite{zhang_frameevent_2023}, CMT-MDNet \cite{wang_visevent_2024}, and CrossEI \cite{chen_crossei_2025}, incorporate cross-attention mechanisms to integrate multimodal features. With the success of visual prompt tuning (VPT) technique, RGB-X tracking has also explored incorporating auxiliary prompts into the RGB ViT backbone \cite{hong_onetracker_2024, wu_singlemodel_2024}. More recently, dual-ViT architectures, such as BAT \cite{chen_bidirectional_2024}, MGNet \cite{zhang_mgnet_2025}, and SMSTracker \cite{chan_smstracker_2025}, enable bidirectional interaction to better exploit complementary cross-modal information.

Although ANNs are highly effective in extracting spatial correlations and global contextual information in RGB images, they are less effective for the sparse spatial structures and rich temporal dynamics of event data. To adapt to the frame-based architecture of ANNs, event streams are often converted into event frames aligned with the RGB frame rate. This results in the loss of fine-grained temporal information. In contrast, spiking neural networks (SNNs) offer a more natural framework for temporal information processing through their intrinsic neural dynamics. This has led to an increasing interest in creating hybrid neural networks (HNNs) that integrate ANNs and SNNs for more effective RGB-Event processing \cite{cao_eventdriven_2023, xiao_respike_2024, wang_sstformer_2025, li_hdiformer_2024}. However, their application to object tracking remains relatively limited. MMHT \cite{sun_reliable_2024} represents the first attempt at HNNs on VOT tasks. This model employs ResNet-based backbones and transformer-based fusion modules for ANN-SNN feature fusion, while its performance is suboptimal. More recently, SNNPTrack \cite{ji_snnptrack_2025} introduces an event prompt using SNNs into a transformer-based RGB tracker. Despite these efforts, the tracking precision of current HNN-based approaches still lags behind the mainstream ANN-based models.

Given the discrepancy between RGB-Event modalities and ANN-SNN paradigms, effective fusion strategies are crucial. As shown in \cref{fig:intro-dual-branch}, similar to ANN-based models, middle fusion \cite{sun_reliable_2024, wang_sstformer_2025} and visual prompt \cite{ji_snnptrack_2025} have been explored. Other works investigate initializing SNN states with ANN features \cite{aydin_hybrid_2024} or employing interactive fusion during feature extraction \cite{xiao_respike_2024, li_hdiformer_2024}. However, most methods rely on heavy attention-based fusion, which leads to high computational costs. Furthermore, these models lack interpretability in how RGB and event information are jointly utilized. In recent years, deep unfolding \cite{gregor_learning_2010, monga_algorithm_2021} has emerged as a model-based paradigm for interpretable multimodal fusion \cite{deng_deep_2021, marivani_designing_2022a, li_learning_2023}. This strategy formulates cross-modal relationships as an optimization problem. Iterative algorithms, such as the iterative shrinkage-thresholding algorithm (ISTA) \cite{daubechies_iterative_2004}, can be unfolded into lightweight network modules to solve this task. This approach bridges principled optimization with efficient deep architectures. However, most existing methods rely on convolutional neural networks (CNNs) and focus on pixel-level fusion. Their application to hybrid ANN-SNN networks and transformer-based frameworks remains largely unexplored. Inspired by these advances, we design a novel ISTA adapter that enables cross-modal interaction between ANN and SNN transformer encoders guided by sparse coding principles.

In this study, we propose \textbf{ISTASTrack}, an \textbf{A}NN-\textbf{S}NN hybrid \textbf{Track}er equipped with \textbf{ISTA} adapters for RGB-Event tracking. To the best of our knowledge, it is the first transformer-based ANN-SNN hybrid framework tailored to multimodal VOT, giving us valuable insights into hybrid network design and cross-paradigm feature fusion. The hybrid architecture incorporates a vision transformer branch to extract structural features from RGB inputs and a spiking transformer branch to exploit the spatio-temporal dynamics inherent in event data. To effectively leverage the complementary features and bridge the disparity between ANN and SNN branches, we design a lightweight ISTA adapter as the core mechanism for cross-paradigm interaction. Specifically, we formulate ANN-SNN interaction as a structured sparse coding problem and solve it through bidirectional unfolded ISTA blocks between the two branches. This formulation provides a principled structural prior that progressively transfers complementary information between modalities through a structured latent representation. Furthermore, a temporal downsampling attention module (TDA) is integrated into the ISTA adapter to enhance temporal feature fusion by aligning multi-step SNN features with single-step ANN features in the latent space. Experimental results have demonstrated that our ISTASTrack consistently outperforms state-of-the-art trackers across multiple datasets, such as FE240hz, VisEvent, COESOT, and FELT, validating the effectiveness and efficiency of hybrid architecture and feature fusion strategies. 

The main contributions of this work are summarized below:

1) We propose the first transformer-based ANN-SNN hybrid framework for RGB-Event tracking. This model effectively leverages contextual and spatial information from RGB data and dynamic temporal cues from event data, achieving state-of-the-art performance with high efficiency.

2) A novel ISTA adapter is systematically developed based on the sparse representation theory. This lightweight adapter enables bidirectional interactive feature adaptation and fusion between heterogeneous features. In addition, a TDA module is incorporated to facilitate semantic temporal alignment and fusion between the SNN and ANN features.

3) ISTASTrack achieves state-of-the-art accuracy with low computational cost on extensive RGB-Event tracking benchmarks, including FE240hz, VisEvent, COESOT, and FELT. The comprehensive analyses further provide insightful guidance for the design of effective and efficient hybrid networks.

\section{Related Works}
\label{sec:review}

\subsection{ANN-based RGB-Event Object Tracking}
\label{subsec:review_VOT}

Visual object tracking (VOT) aims to locate a target in each video frame by matching a given template from the first frame with a search region in subsequent frames. ANN-based VOT methods have evolved from early Siamese neural networks (e.g., SiamFC \cite{bertinetto_fullyconvolutional_2016} and SiamRPN \cite{li_high_2018}) and discriminative learning frameworks (e.g., ATOM \cite{danelljan_atom_2019}, DiMP \cite{bhat_learning_2019}, and PrDiMP \cite{danelljan_probabilistic_2020}) to transformer-based architectures (e.g., STARK \cite{yan_learning_2021}, SwinTrack \cite{chen_swintrack_2021}, AiATrack \cite{gao_aiatrack_2022}, MixFormer\cite{cui_mixformer_2022}, and OSTrack \cite{ye_joint_2022}). The transformer framework has now become the mainstream due to their superior global modeling capacity and robust feature representation.

Early works on RGB-Event tracking extend single-modal frameworks by incorporating multimodal fusion. For example, FENet \cite{zhang_object_2021} introduces cross-domain attention with adaptive weighting to balance modality contributions. AFNet \cite{zhang_frameevent_2023} incorporates alignment and fusion modules guided by event motion cues. CMT-MDNet \cite{wang_visevent_2024} integrates a cross-modality transformer for fusion. More recently, CrossEI \cite{chen_crossei_2025} introduces explicit motion estimation and semantic modulation for scene alignment. However, the performance of these methods remains limited due to the representational constraints of CNN-based backbones.

In recent years, one-stage transformer trackers such as OSTrack \cite{ye_joint_2022} have significantly advanced RGB-Event VOT. CEUTrack \cite{tang_revisiting_2026} proposes the first unified transformer-based framework, enabling joint feature extraction and fusion. Building on this, AMTTrack \cite{wang_longterm_2024} incorporates modern Hopfield layers to enhance associative memory under incomplete multimodal information. To address large modality discrepancies, CRSOT \cite{zhu_crsot_2025} introduces probabilistic feature representations and uncertainty-aware fusion. TENet \cite{shao_tenet_2025a} further designs an event-specific backbone with multi-scale pooling to extract motion-aware features before cross-attention interaction.
%EMTrack \cite{liu_emtrack_2024}

In parallel with intricate fusion module design, another line of research focuses on lightweight adaptation. Approaches such as ViPT \cite{zhu_visual_2023}, OneTracker \cite{hong_onetracker_2024}, and UnTrack \cite{wu_singlemodel_2024} for RGB-X tracking employ the visual prompters between transformer blocks. They design adapters to transform features from the X-modality into RGB prompts as prior. More recently, dual-ViT architectures introduce bidirectional adapters \cite{chen_bidirectional_2024} or dedicated fusion modules \cite{zhang_mgnet_2025, chan_smstracker_2025} to leverage complementary features. Inspired by such design, we introduce interactive ISTA adapters for hybrid architecture.

\subsection{Hybrid Neural Networks}
\label{subsec:review_HNN}

Inspired by complementary learning mechanisms \cite{oreilly_complementary_2014, kumaran_what_2016} observed in biological brains, hybrid neural networks (HNNs) that integrate ANNs and SNNs have demonstrated advantages in perception, cognition, and learning \cite{zhao_framework_2022, liu_advancing_2024, wu_adaptive_2024, shi_hybrid_2025}. Recent RGB-Event studies have explored HNNs for action recognition \cite{xiao_respike_2024}, video restoration \cite{cao_eventdriven_2023}, object detection \cite{wang_sstformer_2025, li_hdiformer_2024}, and tracking \cite{zhang_spiking_2022, sun_reliable_2024}. 

%CH-HNN \cite{shi_hybrid_2025} mitigates catastrophic forgetting in continual learning by leveraging corticohippocampal-inspired circuits to enable energy-efficient, dual-memory representations. Additionally, hybrid spatiotemporal neural networks (HSTNNs) \cite{wu_adaptive_2024} improve adaptability across different performance metrics by flexibly adjusting the proportion of artificial and spiking neurons.

%This requires maintaining information integrity and avoiding computational bottlenecks during modality transitions.  Liu \etal \cite{liu_motion-oriented_2024} proposed a temporal-local-spatio spiking transformer to extract motion saliency and designed a motion-guided SNN-CNN hybrid encoder to fuse motion and background features for image deblurring. Another study \cite{zheng_dance_2022} incorporated temporal binding theory from neuroscience into ANN architectures, combining the temporal dynamics of SNNs with ANN-based attention mechanisms to address multi-object binding challenges.

The key challenge of HNNs lies in designing fusion mechanisms that align ANN and SNN features across spatial and temporal domains. Similar to ANN-based paradigms, some methods employ middle-stage fusion after modality-specific feature extraction. For example, Zhao \etal propose hybrid units to flexibly integrate and decouple features from ANNs and SNNs \cite{zhao_framework_2022}. SC-Net \cite{cao_eventdriven_2023} combines a spiking convolutional backbone with a spatial aggregation module. SSTFormer \cite{wang_sstformer_2025} and MMHT \cite{sun_reliable_2024} employ transformer-based modules to fuse ANN and SNN features.

However, treating ANN and SNN feature extraction separately limits the ability of middle-stage fusion to exploit their complementary strengths. Therefore, some studies introduce ANN-SNN interaction mechanisms during feature extraction. For example, STNet \cite{zhang_spiking_2022} utilizes global spatial context from a transformer to adapts membrane thresholds in SNNs. Similarly, Aydin \etal initialize SNN states with a low-frequency auxiliary ANN to mitigate transient degradation \cite{aydin_hybrid_2024}. ReSpike \cite{xiao_respike_2024} integrates multi-scale cross-attention during ResNet feature extraction, while HDI-Former \cite{li_hdiformer_2024} proposes biologically inspired dynamic interactions between transformer encoders for cross-modal information exchange.

Research on HNNs for RGB-Event VOT is still in its early stages. For example, MMHT \cite{sun_reliable_2024} employs simple spike-based AlexNet and ResNet backbones followed by cross-attention fusion. This simple architecture restrict its performance. More recently, SNNPTrack \cite{ji_snnptrack_2025} introduces SNN-based modules to extract temporal cues from events and fuse them into transformer encoders. However, its unidirectional design fails to fully exploit the complementary representations of both modalities. In contrast, our approach leverages transformers to capture rich contextual information. We introduces a bidirectional feature adaptation mechanism to enable more effective interaction between the ANN and SNN branches.

\subsection{Sparse Coding Model for Multimodal Fusion}
\label{subsec:review-ista}

Sparse coding \cite{elad_sparse_2010} is a classical representation learning framework that represents input signals as a sparse linear combination of atoms from an overcomplete dictionary. It offers a framework to capture the underlying structure or essential features of the data using a compact representation. Sparse coding problems are typically solved by iterative optimization algorithms such as the iterative shrinkage-thresholding algorithm (ISTA) \cite{daubechies_iterative_2004}, approximate message passing \cite{donoho_messagepassing_2009}, and alternating direction method of multipliers \cite{boyd_distributed_2011}. With the rise of deep learning, algorithm unfolding (or unrolling) \cite{gregor_learning_2010, monga_algorithm_2021} has emerged as an effective strategy to transform these iterative procedures into interpretable deep network architectures. This model-driven approach has demonstrated success across various computational imaging tasks, including image restoration \cite{wang_event_2020}, super-resolution \cite{yu_learning_2023}, and image reconstruction \cite{liu_sensing_2023}.

%% by leveraging sparsity constraints and structured representations

%approximate message passing (AMP) \cite{donoho_messagepassing_2009}, and alternating direction method of multipliers (ADMM) \cite{boyd_distributed_2011}

In the context of multimodal fusion, sparse coding facilitates the separation of shared and modality-specific features. Many studies design deep sparse coding models to extract task-specific features from each modality. For example, Deng \etal  propose a multimodal convolutional sparse coding model to extract the common information for image restoration and fusion \cite{deng_deep_2021}. They further  formulate multimodal image registration as a disentangled convolutional sparse coding model to separate registration-relevant and irrelevant features across modalities \cite{deng_interpretable_2023}. Li \etal propose to learn sparse and discriminative multimodal features for finger recognition \cite{li_learning_2023}.

%Similarly, Marivani \etal \cite{marivani_designing_2022a} use the method of multipliers to design CNNs.

However, most deep unfolding modules are built upon CNN architectures, focusing mainly on pixel- or patch-level reconstruction. Their integration into transformer-based multimodal fusion frameworks remains unexplored. To this end, we propose a novel ISTA adapter that introduces interaction mechanisms between ANN-SNN transformer encoders. This adapter enables explicit cross-modal feature mapping in the transform domain guided by sparse coding principles.

\section{Methodology}
\label{sec:methodology}

In this section, we introduce the proposed ISTASTrack framework for RGB-Event tracking. We begin with an overview of the hybrid architecture in \cref{subsec:method-overview}. Next, \cref{subsec:method-backbone} presents the transformer-based ANN and SNN backbones for feature extraction. In \cref{subsec:method-ista-adapt}, we design an ISTA adapter based on sparse representation and algorithm unfolding to promote heterogeneous feature adaptation across transformer encoders. \cref{subsec:method-downta} then describes a temporal downsampling attention module to align multi-step SNN features with single-step ANN features. Finally, \cref{subsec:method-energy} provides an analytical study of network computational cost.

\subsection{Overview of ISTATrack}
\label{subsec:method-overview}

Single visual object tracking (VOT) begins with a known target location in the first frame of a video and aims to localize the target in subsequent frames. Specifically, given the initial bounding box $B_0$ of the target object in the first RGB frame $I_0$, a tracking model $\mathcal{T}$ predicts the position and size of the target $\mathcal{B}_i = (c_{x_i}, c_{y_i}, w_i, h_i)$ in the following frames $I_i$. At each step, the model receives a template image $\bm{Z}_{I_0} \in \mathbb{R}^{3\times H_1 \times W_1}$ centered on the initial target and a search image $\bm{X}_{I_i} \in \mathbb{R}^{3\times H_2 \times W_2}$ cropped around the previous prediction. The template is typically half the size of the search image in both height and width, where $H_1 = H_2 / 2$ and $W_1 = W_2 / 2$.

In the RGB-Event system, two asynchronous data streams are generated. A sequence of RGB frames ${\bm I_i}$ captured at time steps $t_i$, and a continuous stream of events $\{e_j, j=1,2, \cdots, J\}$. Events that occur between two consecutive frames at $(t_i, t_{i+1})$ are accumulated and encoded into an event representation $\bm{E}_i$. In this work, we construct $\bm{E}_i$ as a stack of $T$ event frames for consistency with previous studies \cite{xu_revisiting_2022, zhang_frameevent_2023, wang_longterm_2024}. This results in a temporally aligned input pair $(\bm I_i, \bm{E}_i)$ for each time step. Based on this, the RGB tracking pipeline can be extended to support dual-modality inputs.

\begin{figure*}[tb]
\centering 
{\includegraphics[width=0.9\textwidth]{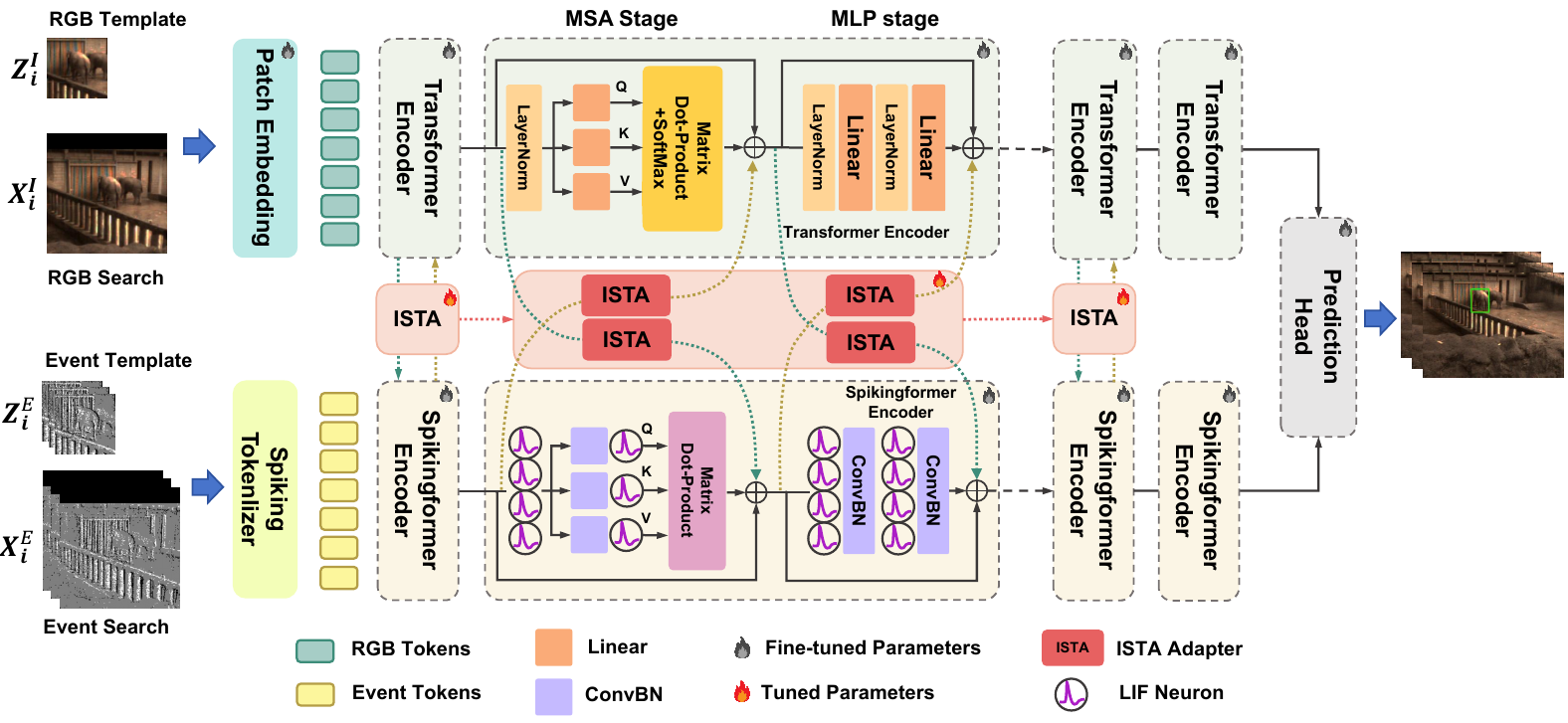}}
\caption{Overview of the ISTATrack architecture. The hybrid network incorporates a vision transformer branch for RGB inputs and a spiking transformer branch for event data. ISTA adapters are incorporated in the first $K$ transformer encoder layers to enable bidirectional feature adaptation and fusion. During training, ISTA adapters are learned from scratch, while the pretrained backbone parameters are fine-tuned with a smaller learning rate.}
\label{fig:istastrack-arch}        
\end{figure*}

As shown in \cref{fig:istastrack-arch}, the proposed RGB-Event tracking framework takes four inputs at each step, i.e., the RGB template $\bm{Z}_{I_0}$ and search image $\bm{X}_{I_i}$, together with the corresponding event template $\bm{Z}_{E_0} \in \mathbb{R}^{T \times 3 \times H_1 \times W_1}$ and search image $\bm{X}_{E_i} \in \mathbb{R}^{T \times 3 \times H_2 \times W_2}$. The predicted bounding box $\mathcal{B}_i$ in frame $i$ is given by
\begin{equation}
    \mathcal{B}_i = \mathcal{T}(\bm Z_{I_0}, \bm Z_{E_0}, \bm X_{I_i}, \bm X_{E_i}, \mathcal{B}_0).
\end{equation}
% where ${B}_i = (c_{x_i}, c_{y_i}, w_i, h_i)$ denotes the center position and size of the predicted object.

The architecture of ISTASTrack is depicted in \cref{fig:istastrack-arch}. The hybrid network employs a dual-branch structure. The ANN branch processes RGB inputs through stacked transformer encoders and the SNN processes event inputs with spiking transformer encoders. Details are provided in \cref{subsec:method-backbone}. To enable cross-modal feature interaction, four ISTA adapters are integrated into each layer of the ANN-SNN encoders. Specifically, two adapters transfer features from ANN to SNN and two from SNN to ANN, with each pair operating at different processing stages. Within the SNN-to-ANN adapter, a temporal downsampling attention module aligns multi-step SNN features with single-step ANN features. Finally, the fused features from both branches are added and forwarded to the prediction head to estimate the object center and size within the search region.

\subsection{ANN and SNN Feature Extraction}
\label{subsec:method-backbone}

\subsubsection{ANN Backbone}
For the RGB branch, we utilize the vision transformer (ViT) backbone consisting of 12 encoder layers from OSTrack \cite{ye_joint_2022}. The RGB template and search images are first transformed into patch embeddings separately, followed by the addition of learnable positional encodings. The resulting tokens are concatenated to form $\bm x^{I} \in \mathbb{R}^{M \times N}$, where $M$ denotes the embedding dimension and $N$ is the total number of tokens. These tokens are subsequently processed by a stack of transformer encoder blocks. Each encoder consists of a multi-head self-attention (MSA) module and a multi-layer perceptron (MLP) module, along with layer normalization and residual connections to enhance training stability. 

%In our implementation, we employ a pretrained model consisting of 12 encoder layers.

\subsubsection{SNN Backbone}

In symmetry with the ANN backbone, we apply a modified version of SpikingFormer \cite{zhou_spikingformer_2023} for the event branch to model global dependencies through spiking self-attention. SpikingFormer is a spike-driven variant of the ViT that significantly reduces energy consumption via the event-driven computing nature of SNNs. It comprises a spiking tokenizer for patch embedding and a series of spiking transformer blocks. The architecture leverages an event-driven residual learning framework based on Leaky Integrate-and-Fire (LIF) neurons, where inputs are first converted into binary spike events (0/1) before linear operations to minimize non-spike computation. The integration of LIF with convolutional layers and batch normalization (ConvBN) contributes to energy efficiency. As illustrated in \cref{fig:istastrack-arch}, the SNN encoder differs from the standard ViT architecture. In self-attention computation, event embeddings are processed by LIF neurons and ConvBN layers, followed by another LIF stage to generate spiking queries (Q), keys (K), and values (V). Attention weights are computed through dot-product operations omitting the softmax function due to the non-negativity of spikes. The MLP block is replaced with a lightweight module consisting of two LIF-based ConvBN layers.

The original SpikingFormer is designed for single-image classification. To adapt it for object tracking and ensure architectural symmetry with the ANN backbone, we introduce two modifications. First, within the spiking tokenizer, the ConvBN layers are shared to process both template and search images, while separate sets of LIF neurons are used for each to maintain distinct neuronal states. Second, the original use of 2D convolution in the MLP stage is not suitable for the 1D token sequences generated by concatenating template and search embeddings. We replace it with 1D convolution, allowing joint processing of the combined tokens $\bm x^{E} \in \mathbb{R}^{T\times M \times N}$. 

It is also worth noting that the pretrained SpikingFormer consists of 8 encoder layers, which is inconsistent with the 12-layer ANN backbone. We align the 8 spiking layers with the first 8 ANN layers and enable cross-modal feature interaction between them. Results in \cref{subsec:result-ablation} demonstrate that this depth asymmetry between the two branches does not compromise tracking performance, while ensuring high computational efficiency.

\subsection{ISTA Adapter for Interactive Feature Fusion}
\label{subsec:method-ista-adapt}

\subsubsection{Modeling Cross-Modality Relations based on Sparse Representation}

In the context of sparse representation, a signal can be represented as a linear combination of a small number of atoms from an overcomplete dictionary. For feature embeddings in the transformer encoder, we denote RGB-ANN features as $\bm x^I \in \mathbb{R}^{M\times N}$ and Event-SNN features as $\bm x^E \in \mathbb{R}^{T\times M\times N}$. When $T=1$, the event representation $\bm x^E$ reduces to the same shape as $\bm x^I \in \mathbb{R}^{M\times N}$. Under this condition, we assume that both RGB and event features can be represented using separate dictionaries, i.e., $\bm x^I = \bm D_I \bm a^I$ and $\bm x^E = \bm D_E \bm a^E$. Here, $\bm D_I, \bm D_E \in \mathbb{R}^{M\times D}$ are learned dictionaries, and $\bm a^I, \bm a^E \in \mathbb{R}^{D\times N}$ are their corresponding sparse codes. When $T>1$, the same dictionary is applied across all time steps, and then a temporal downsampling attention module is introduced to aggregate the $T$-step sparse codes into a single-step representation (see \cref{subsec:method-downta}).

To transfer complementary information from the RGB to the event domain, we transform the ANN feature embedding $\bm x^I$ into its SNN counterpart $\bm x^{E^\prime}$. Although RGB and event data have distinct sensing characteristics, they observe the same physical scene and thus share correlated structural information. We therefore assume that their feature embeddings can be effectively related through a structured latent space. Specifically,
\begin{equation}
\begin{aligned}
    \bm x^I &= \bm D_I \bm a^I, \\
    \bm x^{E^\prime} &= \bm D_E^\prime \bm a^{E^\prime}, \\
    \bm a^I &\approx \bm a^{E^\prime} = \bm a^{I\rightarrow E}.
\end{aligned}
\end{equation}
Similarly, information can be transferred from the Event-SNN to the RGB-ANN branch through a corresponding sparse representation $\bm a^{E\rightarrow I}$,
\begin{equation}
\begin{aligned}
    \bm x^E &= \bm D_E \bm a^E, \\
    \bm x^{I^\prime} &= \bm D_I^\prime \bm a^{I^\prime}, \\
    \bm a^E &\approx\bm a^{I^\prime}=\bm a^{E\rightarrow I}.
\end{aligned}
\end{equation}
Note that this formulation does not require identical sparse codes across modalities. Instead, the decoupled bidirectional sparse codes translate complementary information into latent representations usable by the other branch. % These sparse representations provide a structured latent space that bridges cross-modal feature transfer between the two network branches.

If suitable dictionaries $\bm D_I$ and $\bm D_E$ are available, cross-modal feature adaptation can be formulated as a sparse coding problem, as follows
\begin{equation}
\begin{aligned}
    \min_{\bm a^{I\rightarrow E}} & \|\bm x^{I}-\bm D_I \bm a^{I\rightarrow E}\|_2^2 + \lambda\|\bm a^{I\rightarrow E}\|_1, \\
    \min_{\bm a_i^{E\rightarrow I}} & \|\bm x^{E}-\bm D_E \bm a^{E\rightarrow I}\|_2^2 + \lambda\|\bm a^{E\rightarrow I}\|_1
\end{aligned},
\label{eq:sp-model}
\end{equation}
where $\|\cdot\|_1$ denotes the $\ell_1$ norm. Once the optimal sparse code $\bm a^{I\rightarrow E^*}$ and $\bm a^{E\rightarrow I^*}$ are obtained, the corresponding transformed feature embeddings can be constructed by
\begin{equation}
\begin{aligned}
    \bm x^{E^\prime} &= \bm D_E^\prime \bm a^{I\rightarrow E^*}, \\
    \bm x^{I^\prime} &= \bm D_I^\prime \bm a^{E\rightarrow I^*}.
\end{aligned}
    \label{eq:mapped-feat}
\end{equation}
In fact, \cref{eq:sp-model} defines a LASSO problem that can be solved using the iterative shrinkage-thresholding algorithm (ISTA) \cite{daubechies_iterative_2004}. 
The LASSO-based formulation provides a principled mechanism for structured cross-modal feature transfer and information decoupling. Specifically, the reconstruction term encourages the model to discover latent primitives that capture the underlying scene structure, while the $\ell_1$ penalty promotes sparsity and suppresses modality-specific noise. In this way, the sparse latent representation selectively preserves informative structures while filtering redundant responses.

\subsubsection{ISTA Adapter}

Based on the ISTA algorithm for solving \cref{eq:sp-model}, and omitting modality subscripts for clarity, the sparse coefficients $\bm a$ for a given signal $\bm x$ are iteratively updated following
\begin{equation}
\bm a^{k} = h_{\bm \theta}(\bm a^{k-1} + \bm P(\bm x-\bm D \bm a^{k-1})),
\label{eq:org-ista}
\end{equation}
where $k=1,\cdots, K$ denotes the number of iterations, and $h_{\bm \theta}(\bm x)=sign(\bm x)(\bm x- \bm \theta)_+$ is the soft thresholding function with $\theta$ representing the learnable threshold vector. Here we treat $\bm P$ and $\bm D$ as independent learnable matrices to increase the capacity of the model. The ISTA algorithm iteratively optimizes $\bm a $ with \cref{eq:org-ista} until the convergence criteria are satisfied. 

\begin{figure*}[tb]
\centering 
{\includegraphics[width=0.8\textwidth]{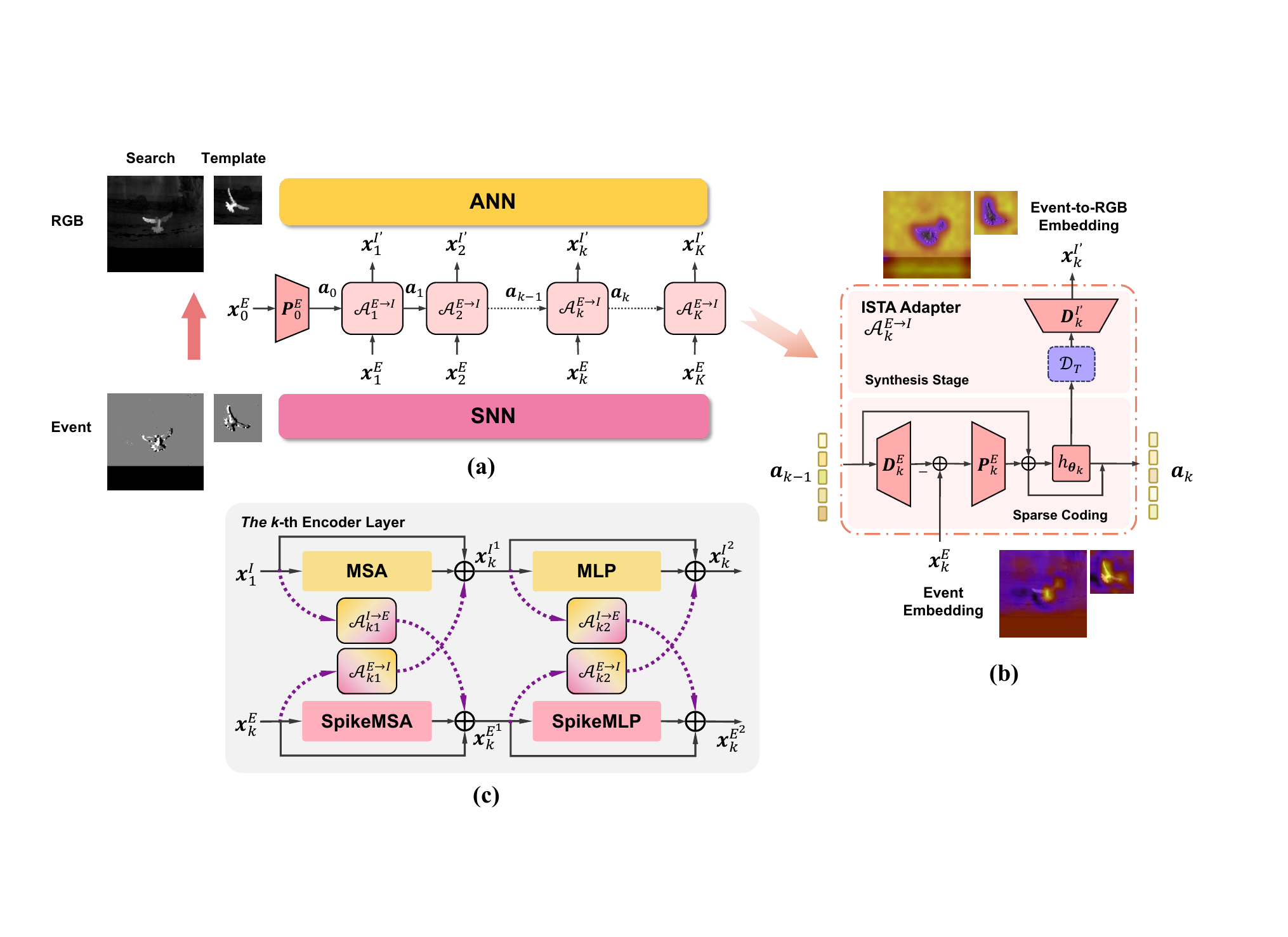}}
\caption{
ISTA adapters for bidirectional feature adaptation and fusion. 
(a) ISTA adapters $\mathcal{A}^{E \rightarrow I}$ from SNN to ANN. The initial sparse codes $\bm a_0$ are generated using the input tokens $\bm x_0^E$. 
(b) Structure of the $k$-th ISTA adapter, composed of a sparse coding stage and a synthesis stage, where $\mathcal{D}_T$ is an optional TDA module for multi-step SNNs. 
(c) Four adapters are integrated into the $k$-th transformer encoder layer at both the MSA and MLP stages.}
\label{fig:ista-adapter}        
\end{figure*}

Following the algorithm unfolding strategy \cite{monga_algorithm_2021}, we construct a stack of ISTA adapters by unrolling several ISTA iterations into learnable modules. The $k$-th iteration is mapped to the $k$-th ISTA adapter, as illustrated in \cref{fig:ista-adapter}(a). 

As shown in \cref{fig:ista-adapter}(b), the operation of an ISTA adapter $\mathcal{A}^{E \rightarrow I}$ from the Event-SNN to the RGB-ANN branch can be formulated as
\begin{equation}
    \begin{aligned}
        \bm a_k &= \bm a_{k-1} + \bm P^E_k(\bm x_k^E-\bm D^E_k \bm a_{k-1}), \\
        \bm a_k &= \frac{1}{2} \left( h_{\bm\theta_k}(\bm a_k) + \bm a_k \right), \\
        \bm a_k &= \mathcal{D}_{T}(\bm a_k) \qquad \text{optional if}\quad T>1,\\
        \bm x_k^{I^\prime} &= \bm D_k^{I^\prime} \bm a_k,
    \end{aligned}
    \label{eq:ista-adapter}
\end{equation}
where the first three equations correspond to the sparse coding stage and the last one represents the feature synthesis stage. In the sparse coding stage, distinct $\bm D_k$, $\bm P_k$, and $\bm \theta_k$ are used at each iteration, where dictionaries $\bm D_k$, $\bm P_k$ are implemented as linear layers and the threshold $\bm \theta_k \in \mathbb{R}^D$ is a learnable vector for each latent element. A skip connection around the soft-thresholding activation is used to promote stable training. When $T>1$, the TDA module $\mathcal{D}_T$ aggregates multi-step sparse codes into a single-step representation. In the feature synthesis stage, the sparse codes are projected by a synthesis dictionary $\bm D_k^{I^\prime}$ to reconstruct the feature embeddings for the RGB-ANN branch. A similar procedure can be applied to map features from the RGB-ANN branch to the Event-SNN branch.

The optimization process is initialized by computing the initial sparse code as $\bm a_0 = \bm P_0 \bm x_0$. Here, $\bm x_0$ denotes the input feature embedding before entering the transformer encoder and $P_0$ is a linear layer, as depicted in \cref{fig:ista-adapter}(a). Each ISTA adapter propagates its output sparse code to the subsequent layer from layer $k$ to $(k+1)$. This enables iterative refinement of the sparse representation across transformer encoder layers. 

Conceptually, the ISTA adapter acts as a structured sparse information bottleneck for complementary information exchange between the RGB-ANN and Event-SNN branches. These sparse representations selectively transfer shared structural cues and suppress modality-specific noise. This process preserves semantic structures essential for tracking. Meanwhile, the dictionary atoms serve as basis elements that capture fundamental visual primitives. Although the introduction of learnable neural components can reduce strict analytical interpretability and introduce partial black-box behavior \cite{shlezinger_deep_2025}, the resulting representations largely remain consistent with sparse coding principles, as illustrated by the visualization of ISTA features in \cref{subsec:result-vis}. 

\subsubsection{Interactive Feature Adaptation}

An individual ISTA adapter enables feature adaptation from one modality to the other. Therefore, two adapters are required for bidirectional interaction between RGB-ANN and Event-SNN branches. Inspired by \cite{chen_bidirectional_2024}, we further embed ISTA adapters into both the MSA and MLP stages, resulting in four ISTA adapters for each transformer encoder layer.

As illustrated in \cref{fig:istastrack-arch} and \cref{fig:ista-adapter}(c), in the $k$-th transformer encoder layer, the Event-SNN branch receives complementary information from the RGB-ANN branch through two ISTA adapters. The feature updates are formulated as
\begin{equation}
\begin{aligned}
    \bm x_k^{E^1} &= \bm x_k^E + \mathrm{SpikeMSA}(\bm x_k^E) + \mathcal{A}_{k1}^{I\rightarrow E}(\bm x_k^{I}), \\
    \bm x_k^{E^2} &= \bm x_k^{E^1} + \mathrm{SpikeMLP}(\bm x_k^E) + \mathcal{A}_{k2}^{I\rightarrow E}(\bm x_k^{I^1}),
\end{aligned}
\label{eq:ista-i2e}
\end{equation}
where $\mathcal{A}_{k1}^{I\rightarrow E}(\cdot)$ and $\mathcal{A}_{k2}^{I\rightarrow E}(\cdot)$ denote the two ISTA adapters at the MSA and MLP stages at the $k$-th encoder. Here, $\bm x_k^{I}$ and $\bm x_k^{I^1}$ are the input feature embeddings for the two stages in the RGB-ANN branch, respectively. In addition, $\bm x_k^{E^1}$ and $\bm x_k^{E^2}$ denote the corresponding outputs of the SpikeMSA and SpikeMLP stages in the Event-SNN branch.  

In the reverse direction from Event-SNN to RGB-ANN, the interaction is given by
\begin{equation}
\begin{aligned}
    \bm x_k^{I^1} &= \bm x_k^I + \mathrm{MSA}(\bm x_k^I) + \mathcal{A}_{k1}^{E\rightarrow I}(\bm x_k^{E}), \\
    \bm x_k^{I^2} &= \bm x_k^{I^1} + \mathrm{MLP}(\bm x_k^I) + \mathcal{A}_{k2}^{E\rightarrow I}(\bm x_k^{E^1}).
\end{aligned}
\label{eq:ista-e2i}
\end{equation}
% where $\mathcal{A}^{E\rightarrow I}(\cdot)$ denotes the ISTA adapter projecting features from event to RGB modality.

In this design, the bidirectional unfolded ISTA adapters perform a sequence of non-linear projections. These projections progressively transfer and adapt complementary information between RGB-ANN and Event-SNN branches through a structured latent representation. Detailed analysis of ISTA features can be found in \cref{subsec:result-vis}.

\subsection{Temporal Downsampling Attention for Time Alignment}
\label{subsec:method-downta}

In our framework, the SNN branch produces multi-step features, while the ANN branch operates on single-step features. Aligning such temporally asynchronous representation is essential for effective cross-modal fusion. Most existing studies directly average SNN features over time, neglecting the rich temporal dynamics inherent in event-based data. To address this limitation, we propose a temporal downsampling attention (TDA) module that performs adaptive temporal compression while preserving informative time variation.

\begin{figure}[tb]
\centering 
{\includegraphics[width=0.8\linewidth]{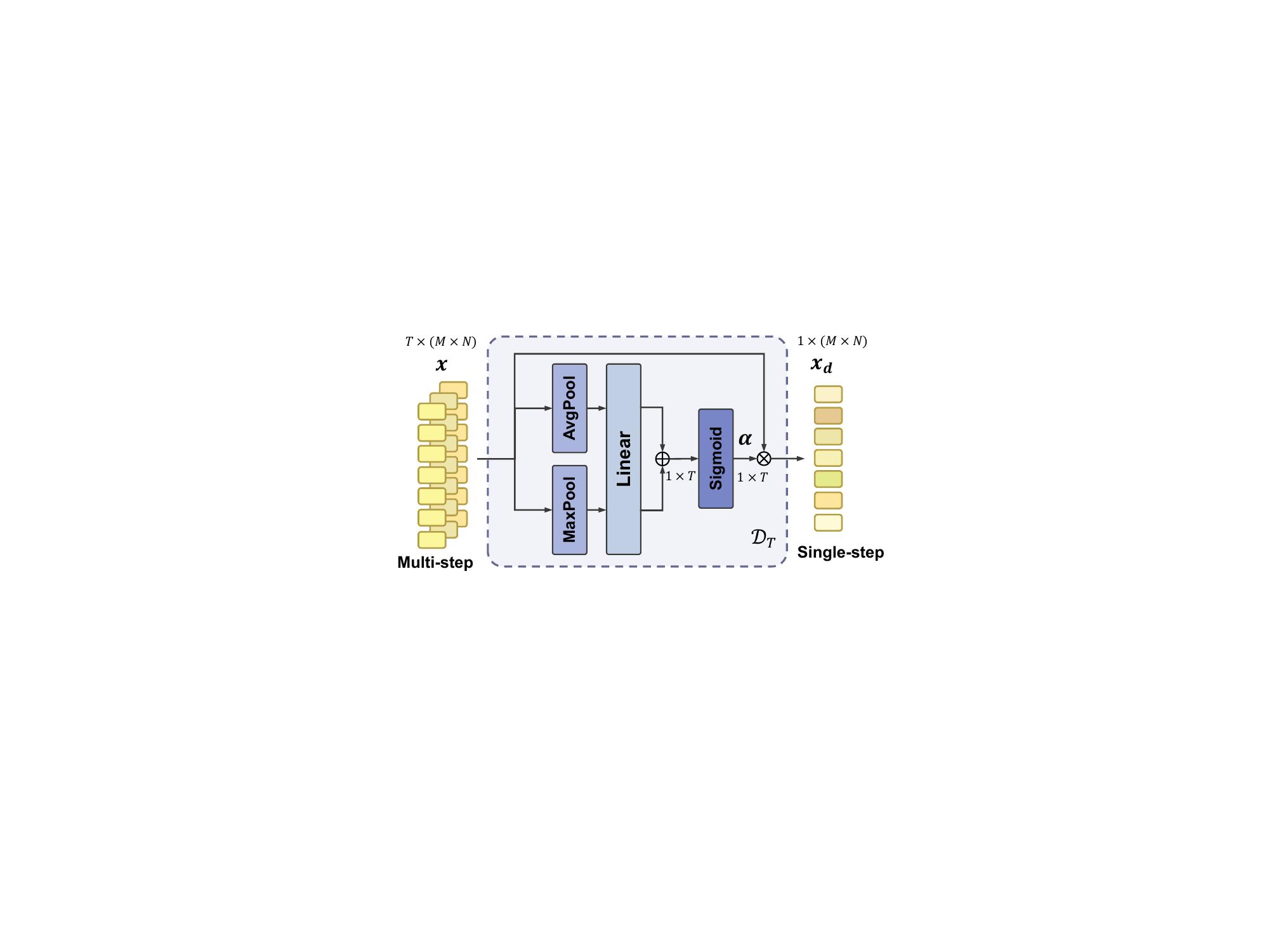}}
\caption{Temporal downsampling attention (TDA) module $\mathcal{D}_T$.}
\label{fig:downta}        
\end{figure}

As depicted in \cref{fig:downta}, the TDA module first receives a sequence of SNN features $\bm x^{E} \in \mathbb{R}^{T \times M \times N}$. Adaptive average pooling and adaptive max pooling are applied separately to the reshaped feature along the temporal dimension, followed by a shared linear operation. The two resulting features are added and then activated by a sigmoid function to produce the temporal attention score vector $\bm \alpha \in \mathbb{R}^{1 \times T}$. This attention score serves as a temporal weighting over the original multi-step feature sequence. The temporally downsampled feature is computed by weighted summation, as follows
\begin{equation}
% \begin{aligned}
    % \bm x^E = \sigma( \text{Linear} ( \text{MaxPool} ( \bm x^{E}) &+ \text{Linear} ( \text{AvgPool} ( \bm x^{E})), \\
    \bm x_d^E = \bm \alpha \bm x^{E},
% \end{aligned}
\end{equation}
where $\sigma(\cdot)$ denotes the sigmoid function and $\bm x_d^E \in \mathbb{R}^{M \times N}$ matches the format of the single-step ANN features.

As shown in \cref{eq:ista-adapter}, The TDA module is embedded within the ISTA adapters from the Event-SNN to the RGB-ANN branch. It aligns the multi-step sparse codes with the single-step ANN representation during feature adaptation. Applying temporal downsampling in the latent sparse space enables efficient semantic temporal fusion while reducing memory cost. The TDA module refines basic temporal averaging through a minimalist design, prioritizing ISTA-based feature adaptation while balancing temporal aggregation and multi-layer integration efficiency.

% As a learned refinement over basic temporal averaging, the TDA maintains a minimalist design to focus learning capacity on the core ISTA-based feature adaptation. This help balance temporal aggregation with multi-layer integration efficiency.

\subsection{Computational Cost Estimation}
\label{subsec:method-energy}

Incorporating an SNN branch offers significant energy efficiency for event data processing. As current GPU/CPU platforms are not optimized for efficient SNN execution, we estimate the computational cost of the network following the widely used approach described in \cite{hu_advancing_2025, zhou_spikingformer_2023}. 

In ANNs, computation is dominated by floating-point multiply-accumulate (MAC) operations. The total number of floating-point operations (FLOPs) is typically used to approximate the computational workload. In contrast, the event-driven and sparse computation paradigm of SNNs is highly efficient for neuromorphic processing \cite{liang_h2learn_2022, qi_ebrainisa_2026}. In SNNs, synaptic operations (SyOPs) refer to the accumulation of postsynaptic inputs triggered by presynaptic spike events. For linear layers following spiking activities, their operations reduce to accumulation (AC) only, since multiplication can be removed due to the binary spike values (0/1). The number of SyOPs is therefore estimated as $fr \times OP_{AC}$, where $fr$ denotes the average neuronal firing rate and $OP_{AC}$ the peak number of accumulation operations in the layer. Nevertheless, SNNs also require non-synaptic MAC operations. Consequently, the computational energy consumption of our hybrid network is estimated by
\begin{equation}
    E = E_{MAC} \times \sum_{i} OP^{i}_{MAC} + E_{AC} \times \sum_{i} (OP^i_{AC} \times fr^i),
\end{equation}
where $i=1\times N$ indexes the network layers, $E_{MAC}=4.6pJ$ and $E_{AC}=0.9pJ$ represent the approximate energy cost of 32-bit floating-point MAC and AC operations on 45nm hardware, respectively \cite{horowitz_11_2014}. A detailed analysis of computational efficiency can be found in \cref{subsec:result-efficiency}.

\section{Experimental Results}
\label{sec:results}

\subsection{Experimental Settings}
\label{subsec:result-settings}

\subsubsection{Datasets}

We utilize four representative RGB-Event tracking datasets, including FE240hz \cite{zhang_frameevent_2023}, COESOT \cite{xu_revisiting_2022}, VisEvent \cite{wang_visevent_2024}, and FELT \cite{wang_longterm_2024}. The four datasets are all collected by DAVIS346 with an image resolution of $346 \times 260$.

The FE240hz dataset covers real-world challenges, such as low lighting and fast object motion. Most scenes are captured by a stationary camera. We use annotations synchronized with the RGB frame rate. The dataset consists of 140K annotated frames, with 101 sequences used for training and 40 for testing.

%extends the FE108 dataset \cite{zhang_object_2021} by introducing more challenging sequences and 

The VisEvent dataset focuses on traffic scenes with pedestrians and vehicles, along with a few indoor scenarios. It comprises approximately 370K frames across 820 sequences. Due to incomplete raw event data, we use 295 sequences for training and 219 for testing.

The COESOT dataset provides a broad range of target categories, including vehicles, pedestrians, and various animals. This benchmark consists of 1,354 aligned sequences, with 827 sequences for training and 527 for testing.

FELT is the largest benchmark for long-term RGB-Event tracking. It contains 742 sequences and approximately 1.6M paired frames. The dataset features complex scenarios like ball sports with small objects and background interference. It is divided into 520 training and 222 testing sequences.

\subsubsection{Training Details}

To train our network, we employ a composite loss function comprising focal loss for classification and a combination of L1 loss and GIoU loss for bounding box regression, as follows
\begin{equation}
\mathcal{L} = \lambda_1 \mathcal{L} _{\text{focal}} + \lambda_2 \mathcal{L} _1 + \lambda_3 \mathcal{L} _{\text{GIoU}},
\end{equation}
where $\lambda_1$, $\lambda_2$, and $\lambda_3$ are trade-off coefficients, empirically set to 1, 5, and 2, respectively, following the setting in \cite{ye_joint_2022}.

For initialization, we load the pretrained weights from OSTrack \cite{ye_joint_2022} for the ANN branch and partially initialize the SNN branch using a pretrained SpikingFormer checkpoint \cite{zhou_spikingformer_2023}, excluding the modified layers. During training, the pretrained parameters are fine-tuned with a smaller learning rate of $1e^{-5}$, while the remaining parameters are trained with a learning rate of $1e^{-4}$. The network is trained for 60 epochs using the Adam optimizer with a weight decay of $1e^{-4}$ and a batch size of 32. The learning rate is reduced to 10\% of its initial value after 48 epochs. The input image sizes are set to $128 \times 128$ for the template and to $256 \times 256$ for the search region. The number of SNN time steps $T$ is set to 3 by default to achieve a balance between accuracy and computational efficiency. The number of layers that integrate ISTA adapters is set to $L=4$ by default, corresponding to the configuration that yields optimal performance (see \cref{subsec:result-ablation}). 

\subsubsection{Evaluation Metrics}

We evaluate tracking performance using several metrics, including success rate (SR), overlap precision (OP$_\text{T}$), precision (PR), and normalized precision (NPR). The SR measures the proportion of frames where the intersection-over-union (IoU) between the predicted and ground truth bounding boxes exceeds a range of thresholds, and we report the area under the curve (AUC) of the SR plot as the final score. OP$_\text{T}$ refers to the SR at a specific IoU threshold T. We report results for OP50 and OP75, corresponding to thresholds of 50\% and 75\%, respectively. PR measures the percentage of frames where the Euclidean distance between the predicted and ground truth bounding box centers falls below a given threshold, which we set to 20 pixels. NPR follows the same definition as PR but normalizes the centers by the ground truth bounding box dimensions. 
%Frames where the target is out of view are excluded from evaluation.

\subsection{Comparison with State-of-the-art Trackers}
\label{subsec:result-compare}

\subsubsection{Quantitative Comparison}

We compare our method against a wide range of state-of-the-art trackers, including single-branch ANN, dual-branch ANN, dual-branch SNN and hybrid ANN-SNN approaches. Results are summarized in \cref{tab:results-compare}. For fair comparison, we retrain most models on the four aforementioned datasets under our unified training settings. If the reported performance cannot be reproduced due to unavailable training details, code, or pretrained checkpoints, we report the results from the original publications. The best performance of ISTASTrack across different SNN time steps is reported, which can be found in \cref{tab:ablation-time-step}.

\begin{table*}[tbp]
\caption{Comparison of tracking performance with other state-of-the-art trackers. \textbf{Bold} values represent the best results for the same dataset.}
    \centering
    \setlength{\tabcolsep}{0.7mm}
    \renewcommand{\arraystretch}{1.4}
    \scriptsize
\begin{tabular}{c|l|c|ccccc|ccccc|ccccc|ccccc}
\hline
\multirow{2}[4]{*}{\textbf{Type}} & \multirow{2}[1]{*}{\textbf{Tracker}} & \multirow{2}[4]{*}{\textbf{Architechture}} & \multicolumn{5}{c|}{\textbf{FE240hz}} & \multicolumn{5}{c|}{\textbf{VisEvent}} & \multicolumn{5}{c|}{\textbf{COESOT}}  & \multicolumn{5}{c}{\textbf{FELT}} \\
\cline{4-23}      &       &       & \textbf{SR} & \textbf{OP50} & \textbf{OP75} & \textbf{PR} & \textbf{NPR} & \textbf{SR} & \textbf{OP50} & \textbf{OP75} & \textbf{PR} & \textbf{NPR} & \textbf{SR} & \textbf{OP50} & \textbf{OP75} & \textbf{PR} & \textbf{NPR} & \textbf{SR} & \textbf{OP50} & \textbf{OP75} & \textbf{PR} & \textbf{NPR} \\
\hline
\multicolumn{1}{c|}{\multirow{7}[2]{*}{\makecell{\textbf{Single-} \\ \textbf{branch} \\ \textbf{ANN}}}} & \textbf{ATOM \cite{danelljan_atom_2019}} & ResNet50 & 38.7  & 49.0  & 19.3  & 57.1  & 43.8  & 42.5  & 50.5  & 28.3  & 58.7  & 61.8  & 56.0  & 66.9  & 49.3  & 66.6  & 73.4  & 37.5  & 43.4  & 31.4  & 44.3  & 63.7  \\
      & \textbf{DiMP \cite{bhat_learning_2019}} & ResNet50 & 57.0  & 72.6  & 27.6  & 87.5  & 63.5  & 46.5  & 55.7  & 32.8  & 64.1  & 65.8  & 59.3  & 70.4  & 51.3  & 69.8  & 76.7  & 45.0  & 53.4  & 38.2  & 53.8  & 72.9  \\
      & \textbf{PrDiMP \cite{danelljan_probabilistic_2020}} & ResNet50 & 55.4  & 70.4  & 26.7  & 84.6  & 61.0  & 46.4  & 54.7  & 34.8  & 62.8  & 65.6  & 58.2  & 68.2  & 52.3  & 67.2  & 74.5  & 45.3  & 53.0  & 39.4  & 53.9  & 72.7  \\
      & \textbf{STARK \cite{yan_learning_2021}} & ResNet+ViT & 52.4  & 64.3  & 24.5  & 83.2  & 59.8  & 44.1  & 52.1  & 34.9  & 58.3  & 60.4  & 58.5  & 67.1  & 53.2  & 66.4  & 73.2  & 47.5  & 54.4  & 42.0  & 58.4  & 73.9  \\
      & \textbf{OSTrack \cite{ye_joint_2022}} & ViT   & 58.6  & 72.5  & 32.3  & 88.8  & 64.7  & 54.4  & 66.0  & 48.0  & 70.8  & 76.7  & 61.6  & 70.4  & 58.1  & 69.5  & 76.3  & 46.2  & 53.8  & 41.5  & 54.8  & 73.6  \\
      & \textbf{AiATrack \cite{gao_aiatrack_2022}} & ViT   & 55.8  & 70.7  & 25.6  & 85.9  & 60.3  & 45.4  & 54.7  & 33.3  & 61.4  & 64.8  & 61.5  & 72.2  & 55.3  & 71.1  & 77.2  & 48.4  & 55.7  & 43.2  & 58.5  & 75.0  \\
      & \textbf{CEUTrack \cite{xu_revisiting_2022}} & Unified ViT & 55.6  & -     & -     & 84.5  & -     & 54.4  & 67.0  & 46.6  & 71.2  & 75.9   & 63.9  & 74.4  & 62.6  & 74.3  & 80.5  & 46.7  & 54.7  & 43.0  & 55.5  & 74.6  \\ 
\hline
\multicolumn{1}{c|}{\multirow{7}[4]{*}{\makecell{\textbf{Dual-} \\ \textbf{branch} \\ \textbf{ANN}}}} & \textbf{FENet \cite{zhang_object_2021}} & Dual ResNet18 & 55.6  & -     & -     & 84.3  & -     & 41.0  & 48.6  & 32.9  & 54.5  & 58.7  & 50.3  & 56.0  & 39.4  & 52.7  & 61.2  & 43.8  & 50.8  & 39.3  & 52.0  & 70.8  \\
      & \textbf{AFNet \cite{zhang_frameevent_2023}} & Dual ResNet18 & 57.8  & 73.1  & 28.9  & 87.8  & 63.7  & 44.5  & -     & -     & 59.3  & -     & 50.9  & 55.7  & 41.6  & 54.0  & 62.6  & 36.9  & 40.9  & 28.0  & 41.9  & 61.2  \\
      & \textbf{CrossEI* \cite{chen_crossei_2025}} & Dual ResNet18 & 59.4  & 74.6  & 28.8  & 92.3  & -     & 53.1  & 62.4  & 40.3  & 71.4  & -     & 61.7  & 70.1  & 59.0  & 70.9  & -     & -     & -     & -     & -     & - \\
      & \textbf{TENet \cite{shao_tenet_2025a}} & ViT+Pooler & 60.3  & 76.9  & 35.0  & 88.4  & 67.7  & 59.5  & 71.5  & 57.0  & 74.7  & 82.2  & 70.5  & 80.8  & 71.4  & 80.7  & 87.1  & 49.6  & 57.1  & 47.1  & 58.1  & 77.0  \\
\cline{2-23}      & \textbf{ViPT \cite{zhu_visual_2023}} & \multicolumn{1}{c|}{\multirow{3}[2]{*}{\makecell{{Dual ViT,} \\ {Vision Prompt}}}} & 60.9  & 78.1  & 36.8  & 88.2  & 69.1  & 58.8  & 70.6  & 55.6  & 74.1  & 81.1  & 69.6  & 79.6  & 70.1  & 79.6  & 85.7  & 47.2  & 54.5  & 44.1  & 55.4  & 74.3  \\
      & \textbf{UnTrack \cite{wu_singlemodel_2024}} &       & 62.0  & 79.4  & 36.4  & 90.1  & 69.2  & 56.5  & 67.1  & 51.4  & 71.6  & 77.7  & 71.0  & 80.5  & 71.9  & 80.5  & 86.8  & 48.2  & 55.5  & 45.4  & 56.3  & 75.4  \\
      & \textbf{BAT \cite{cao_bidirectional_2024}} &       & 64.3  & 82.3  & \textbf{41.6} & 91.9  & \textbf{73.1} & 61.3  & 74.0  & 58.7  & \textbf{77.0} & 84.1  & \textbf{71.4} & 80.5  & 72.7  & 80.8  & 86.7  & 48.4  & 55.5  & 45.6  & 56.7  & 75.4  \\
\hline
\multicolumn{1}{c|}{\textbf{SNN}} & \textbf{SpikeFET} & \multicolumn{1}{c|}{Dual SNN ViT} & 60.3  & 76.4  & 38.5  & 85.6  & 66.5  & 60.7  & 71.9  & \textbf{59.4} & 74.8  & 82.5  & 71.3  & 81.1  & 72.4  & 81.0  & 87.1  & 49.1  & 55.6  & 46.5  & 56.2  & 75.6  \\
\hline
\multicolumn{1}{c|}{\multirow{3}[2]{*}{\makecell{\textbf{ANN-} \\ \textbf{SNN}}}} & \textbf{MMHT \cite{sun_reliable_2024}} & \makecell{ResNet18+ \\ AlexSNN}  & -     & -     & -     & -     & -     & 55.1  & 65.9  & 42.8  & 73.3  & -     & 65.8  & 77.6  & 57.0  & 74.0  & -     & -     & -     & -     & -     & - \\
      & \textbf{SNNPTrack \cite{ji_snnptrack_2025}} & ViT+SNN Prompt & -     & -     & -     & -     & -     & 59.8  & -     & -     & 76.9  & -     & 66.8  & -     & -     & 74.8  & -     & -     & -     & -     & -     & - \\
\cline{2-23}      & \textbf{ISTASTrack(Ours)} & ViT+SNN ViT & \textbf{64.6} & \textbf{82.3} & \textbf{41.6} & \textbf{92.8} & \textbf{73.1} & \textbf{61.4} & \textbf{74.1} & 58.4  & 76.8  & \textbf{84.3} & \textbf{71.4} & \textbf{81.8} & \textbf{73.1} & \textbf{81.7} & \textbf{87.8} & \textbf{50.1} & \textbf{57.6} & \textbf{47.6} & \textbf{58.8} & \textbf{77.5} \\
\hline
\end{tabular}%

    \label{tab:results-compare}
\end{table*}

\textbf{Single-branch ANN}. Classic ANN-based trackers originally designed for single-modality tasks can be adapted for RGB-Event tracking by modifying their first layer to process early-fused inputs, including discriminative learning-based models such as ATOM \cite{danelljan_atom_2019}, DiMP \cite{bhat_learning_2019}, and PrDiMP \cite{danelljan_probabilistic_2020}, as well as transformer-based designs like STARK \cite{yan_learning_2021}, OSTrack \cite{ye_joint_2022}, and AiATrack \cite{gao_aiatrack_2022}. CEUTrack \cite{xu_revisiting_2022} further introduces an early fusion module for ViT, specifically tailored for RGB-Event tracking. As shown in \cref{tab:results-compare}, early-fusion ANN trackers remain limited in effectiveness, among which transformer-based models such as OSTrack, AiATrack, and CEUTrack achieve relatively stronger performance.

\textbf{Dual-branch ANN.} This category includes methods explicitly designed for RGB-Event fusion, such as FENet \cite{zhang_object_2021}, AFNet \cite{zhang_frameevent_2023}, and CrossEI \cite{chen_crossei_2025}. Their effectiveness is constrained by relatively simple ResNet-based feature extractors and complex inference pipelines. In contrast, TENet \cite{shao_tenet_2025a} delivers competitive performance by using a ViT backbone for the RGB branch. We also consider dual-modality transformer models based on visual prompting, including ViPT \cite{zhu_visual_2023}, UnTrack \cite{wu_singlemodel_2024}, and BAT \cite{cao_bidirectional_2024}. These frameworks outperform most models, with BAT achieving the second-best overall results. These results highlight the importance of dual-branch transformer architectures for processing different modalities. 

\textbf{Dual-branch SNN.} SpikeFET \cite{yang_fully_2025} achieves performance comparable to TENet. The fully spiking model with dual spike ConvFormer backbones demonstrates the strong representational capability of SNN-based ViTs. This SNN model also exhibits superior energy efficiency, as reported in \cref{tab:results-energy}.

\textbf{ANN-SNN.} As introduced in \cref{sec:introduction}, only two hybrid ANN-SNN models have been proposed for RGB-Event tracking, i.e., MMHT \cite{sun_reliable_2024} and SNNPTrack \cite{ji_snnptrack_2025}. MMHT utilizes simple ResNet and AlexSNN as backbones, resulting in relatively weak performance. By contrast, SNNPTrack employs SNN-based prompts inserted between ViT encoders to guide fusion, demonstrating stronger results. However, its unidirectional design limits the ability to fully exploit the strengths of both modalities. In comparison, our ISTASTrack employs an SNN-ViT backbone for the event branch and integrates bidirectional adapters for enhanced RGB-Event fusion. Therefore, ISTASTrack fully leverages complementary spatio-temporal information, achieving the best overall performance.

\subsubsection{Visual Comparison of Typical Scenes}

\begin{figure*}[tb]
    \centering
    \includegraphics[width=0.88\linewidth]{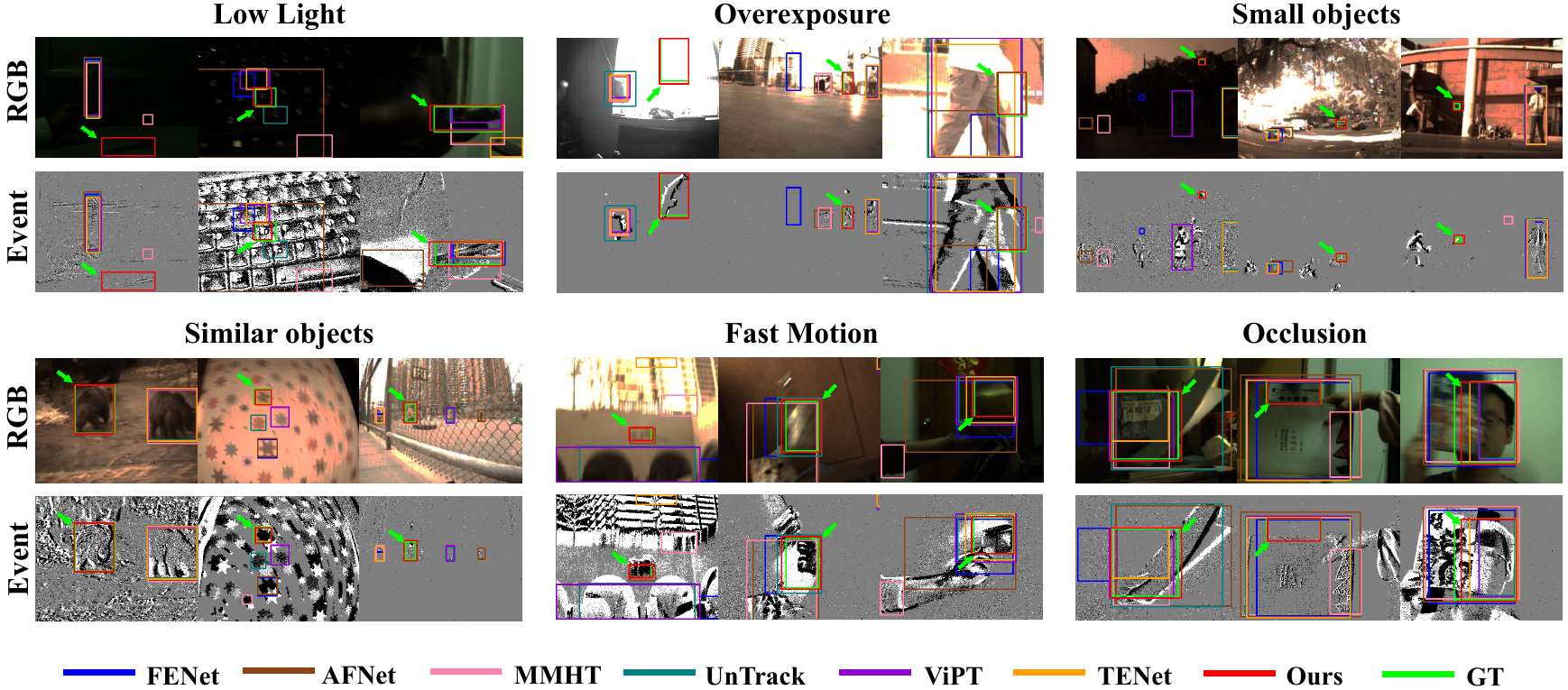}
    \caption{Qualitative comparison of tracking results on challenging long sequences with similar objects and occlusions. \textcolor{green}{Green} boxes and arrows indicate ground truth (GT) targets, and \textcolor{red}{red} boxes represent predictions from ISTASTrack. Our method consistently maintains accurate target tracking despite occlusions and object confusion.}
    \label{fig:results-compare}
\end{figure*}

To further illustrate the comparison, \cref{fig:results-compare} presents the tracking results of representative methods across typical scenarios, including low-light, overexposure, small objects, similar objects, fast motion, and occlusions. In the visualization, green boxes represent the ground truth (GT), while red boxes represent the predictions of our ISTASTrack. The results demonstrate that our network remains robust under challenging lighting, fast motion, and occlusions. Furthermore, the model detects small objects more accurately and effectively distinguishes them from similar distractors.

\begin{figure*}[tb]
    \centering
    \includegraphics[width=0.95\linewidth]{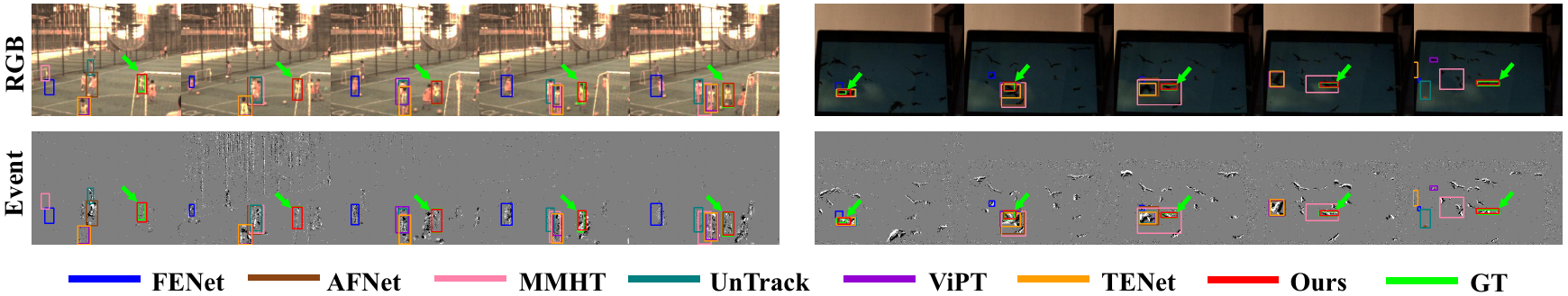}
    \caption{Tracking results on challenging long sequences with similar objects and occlusions. Green boxes and arrows indicate ground truth (GT) targets, and red boxes represent predictions from ISTASTrack. Our method consistently maintains accurate target tracking despite occlusions and object confusion.}
    \label{fig:results-compare-longseq}
\end{figure*}

Furthermore, \cref{fig:results-compare-longseq} presents two challenging sequences that involve similar objects and occlusions. In the left example showing a football game, ISTASTrack successfully tracks the target player despite occlusions from the goal post and other players, whereas other methods lose the target and often drift to distractors. Similarly, in the right example with multiple flying birds, our method reliably re-identifies the correct target after they intermingle. These examples demonstrate the robustness of ISTASTrack in complex scenarios.

\subsubsection{Attribute-based Comparison}

\begin{figure*}[tb]
    \centering
    \includegraphics[width=0.75\linewidth]{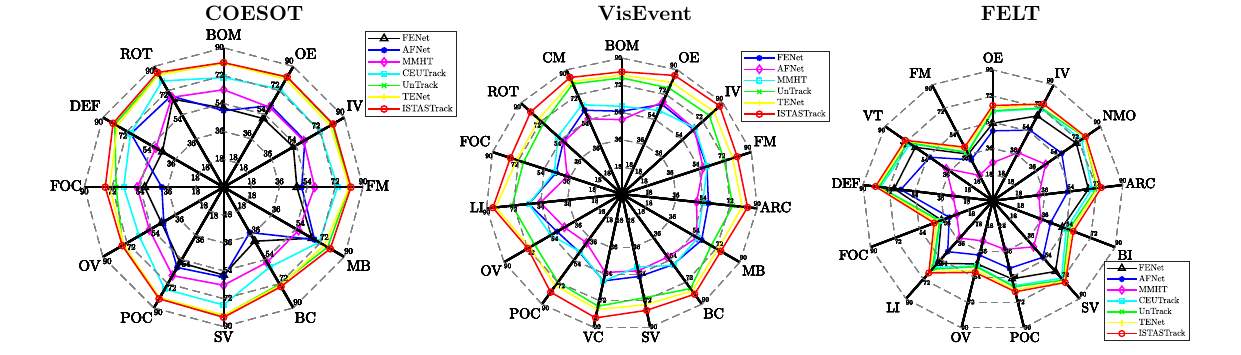}
    \caption{Precision scores across attributes on COESOT, VisEvent, and FELT datasets.}
    \label{fig:results-radar}
\end{figure*}

Since the VisEvent, COESOT, and FELT datasets provide sequence-level attributes, we plot a radar chart in \cref{fig:results-radar} to compare ISTASTrack with representative methods across different scene attributes. Overall, ISTASTrack outperforms other methods in nearly all attributes, particularly on VisEvent. Transformer-based methods with interactive fusion, such as TENet and UnTrack, also show stronger robustness than the remaining baselines. For challenging conditions such as partial occlusion (POC) and full occlusion (FOC), all models experience a drop in performance, yet ISTASTrack demonstrates greater resilience to occlusions. On the most challenging FELT dataset, fast motion (FM), strong background interference (BI), and FOC lead to generally low performance. Our method performs slightly better than the other approaches despite the overall difficulty.

\subsection{Analysis and Visualization of ISTA features} 
\label{subsec:result-vis}
%Visualization of Intermediate Features

\subsubsection{Interpretability of Learned Sparse Codes and Dictionaries}
\begin{figure*}[tb]
    \centering
    \includegraphics[width=0.7\linewidth]{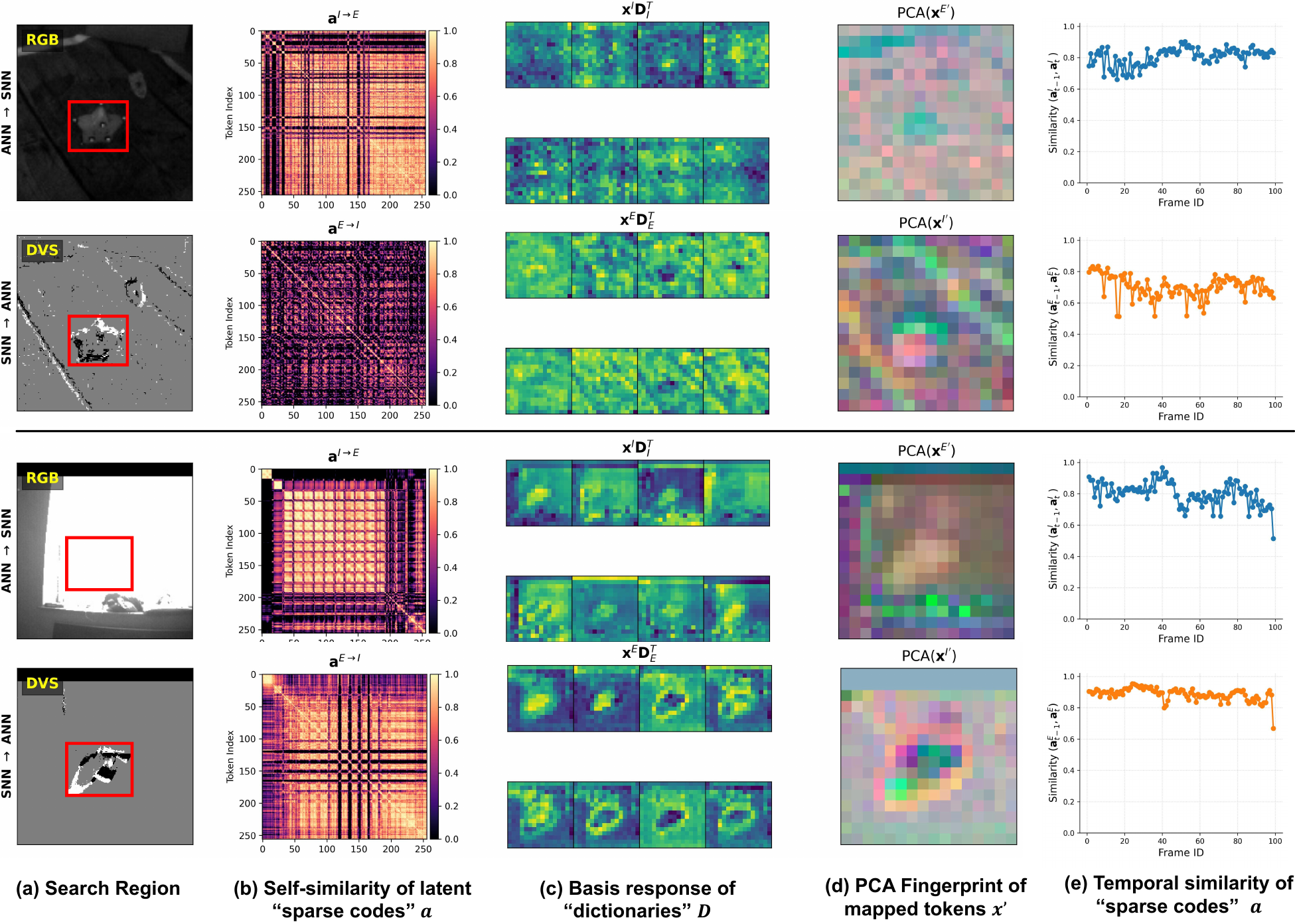}
    \caption{Visualization of learned latent codes and cross-modal features within the bidirectional ISTA adapters. (a) \textbf{Search Region}: Input from RGB and DVS sensors. (b) \textbf{Sparse Code Self-Similarity}: Normalized self-similarity matrices of $\bm a^{I\rightarrow E}$ and $\bm a^{E\rightarrow I}$, showing structured correlation and semantic clustering of latent features. (c) \textbf{Dictionary Basis Responses}: Spatial response maps of learned dictionary atoms $\bm D_I$ and $\bm D_E$. (d) \textbf{PCA Fingerprint}: Feature distributions of ISTA features $\bm x^{E'}$ and $\bm x^{I'}$, illustrating complementary information transfer and cross-modal feature interaction. (e) \textbf{Temporal similarity}: Stability of sparse codes across frames, where fluctuations mainly occur when significant scene changes appear.}
    \label{fig:vis-ista}
\end{figure*}

To assess interpretability, we analyze the latent sparse codes and cross-modal features within the bidirectional ISTA adapters (\cref{fig:vis-ista}).

We first visualize features corresponding to related search regions (see \cref{fig:vis-ista}(a)). The normalized self-similarity of the sparse codes ($S = \bm a \bm a^T$) in \cref{fig:vis-ista}(b) reveals clear block-diagonal structures, indicating selective activation of dictionary atoms and structured sparsity. In \cref{fig:vis-ista}(c), spatial response maps of the learned dictionaries ($\bm x \bm D^T$) show that atoms act as adaptive filters capturing fundamental visual primitives, such as edges and object contours. \cref{fig:vis-ista}(d) presents the PCA fingerprints of ISTA features $\bm x^{E'}$ and $\bm x^{I'}$. The feature distributions illustrate that ISTA adapters transfer complementary structural cues across modalities, emphasizing target-related features while propagating background information between branches. These results demonstrate that the adapters bridge the modality gap while preserving informative structures.

To evaluate temporal stability, we compute the similarity of sparse codes between consecutive frames within the same ISTA block. As shown in \cref{fig:vis-ista}(e), similarity remains around 80\% in stable scenes, indicating robustness to minor temporal variations. SNN features fluctuate more when DVS data contain richer textures due to camera motion (top row), while clean DVS backgrounds yield more stable representations (bottom row). Overall, both branches maintain consistent temporal trends across sequences.

% The $L_1$ regularization, implemented through the soft-shrinkage operation, acts as a principled mechanism for noise suppression and feature selection.
\subsubsection{Effect of Soft-Thresholding} 

\begin{figure*}[tbp]
    \centering
    \includegraphics[width=0.93\linewidth]{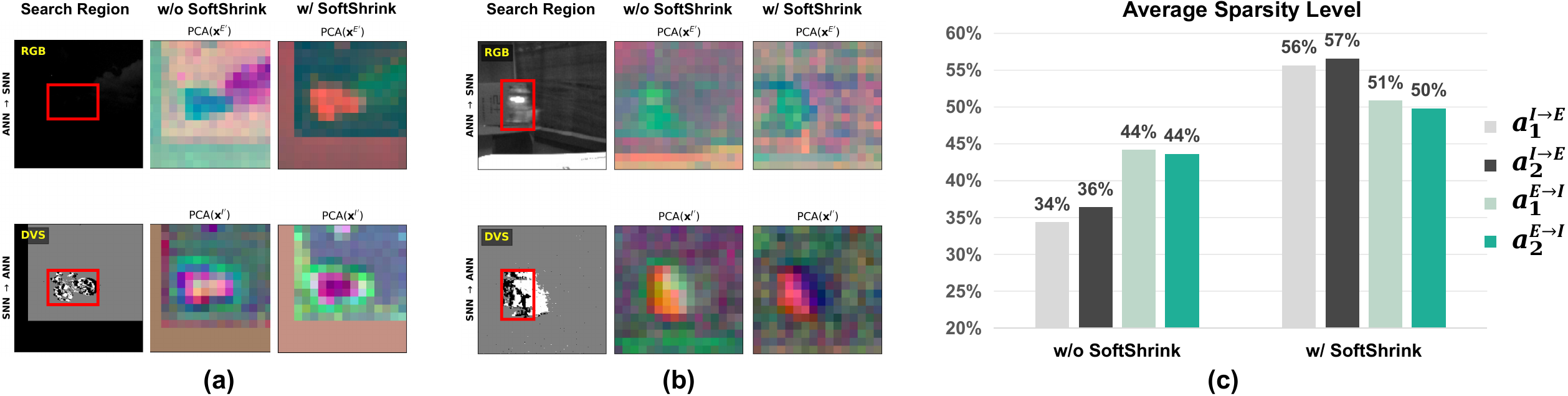}
    \caption{Comparison of ISTA features with (w/ SoftShrink) and without (w/o SoftShrink) the soft-thresholding operation.
(a)(b) \textbf{PCA fingerprint} comparisons of $\bm x^{E'}$ and $\bm x^{I'}$. SoftShrink enhances target-background separability and produces more concentrated features with clearer contours for fast-moving targets.
(c) \textbf{Average sparsity} evaluated by the Top 10\% energy contribution. SoftShrink increases the energy concentration within the top 10\% components of the learned sparse codes $\bm a$, demonstrating enhanced feature sparsity.}
    \label{fig:vis-softshrink}
\end{figure*}

To examine the role of the $\ell_1$ regularization implemented via soft thresholding operation, we compare ISTA features with and without the operation in \cref{fig:vis-softshrink}. As shown in \cref{fig:vis-softshrink}(a), features without soft-thresholding (w/o SoftShrink) exhibit stronger interference from background structures, leading to weaker foreground-background separation in the PCA distributions. In contrast, soft-thresholding suppresses irrelevant responses and improves semantic discrimination. A second example in \cref{fig:vis-softshrink}(b) further shows that the operation produces more concentrated features with clearer contours for fast-moving targets.

To quantify sparsity, we measure the energy concentration of the top 10\% sparse code components. As illustrated in \cref{fig:vis-softshrink}(c), applying soft-thresholding significantly increases the proportion of energy captured by the top components. This indicates that redundant responses are suppressed and the resulting representation becomes more compact and discriminative.

\subsubsection{Cross-Modal Attention Adaptation}
\begin{figure*}[tb]
\centering
    \includegraphics[width=0.7\linewidth]{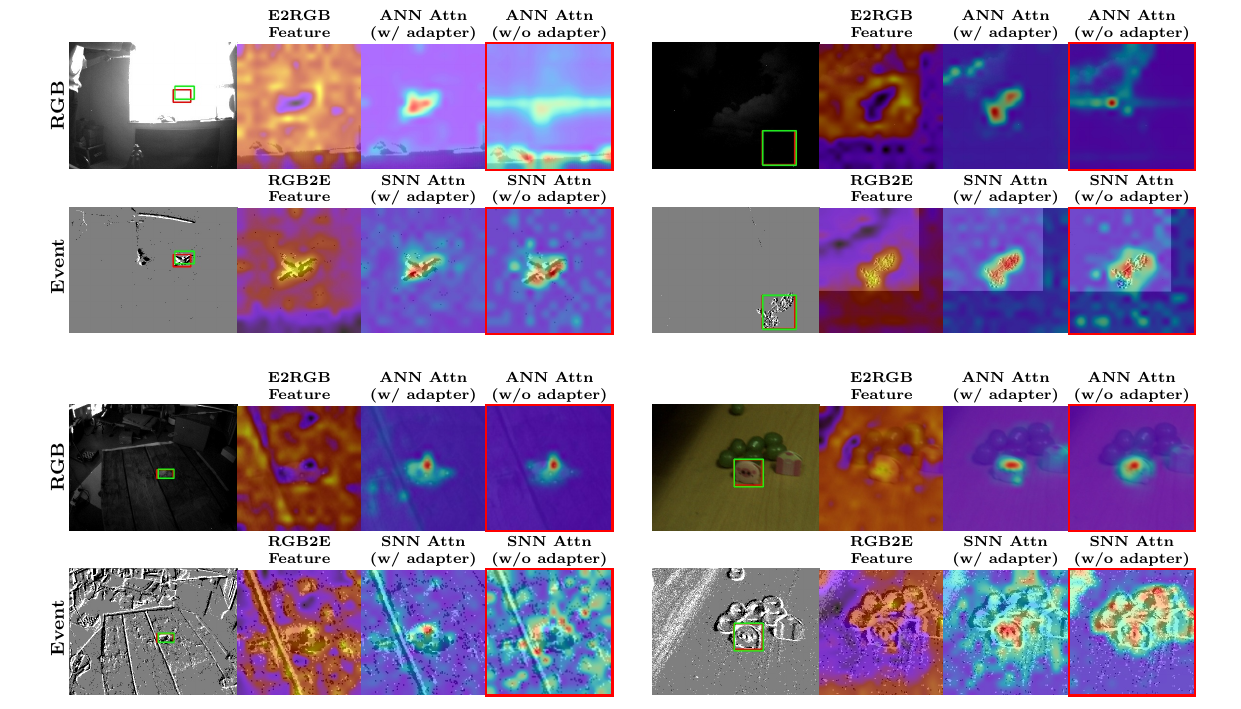}
    \caption{Visualization of ISTA features and attention maps from the first encoder layer. RGB2E and E2RGB represent feature maps transferred by the ISTA adapter from RGB to event and from event to RGB, respectively. We present ANN and SNN attention maps generated by models with or without the ISTA adapter, denoted as Attn w/ adapter and Attn w/o adapter(highlighted with \textcolor{red}{red} bounding boxes), respectively.}
    \label{fig:results-istafunc}
\end{figure*}

To illustrate the role of the ISTA adapter in multimodal feature adaptation, we visualize intermediate representations from the first encoder layer in \cref{fig:results-istafunc}. Specifically, we show time-averaged feature maps of the search region, including features transferred by the ISTA adapters from RGB to event (RGB2E) and from event to RGB (E2RGB), together with the corresponding ANN and SNN attention maps. For comparison, attention maps from a model without ISTA adapters are also provided (\textit{Attn. w/o adapter}), highlighted with red bounding boxes.

We present four challenging examples involving lighting variation, background clutter, and similar objects. In the first row, severe overexposure/underexposure makes the target difficult to identify in the RGB image. This causes the ANN attention without ISTA adapters to miss the target. With the help of ISTA adaptation, the ANN successfully focuses on the target, and the SNN attention also becomes more accurate through the transferred features. In the last row, where scenes contain rich textures, the ANN can localize the target. However, the SNN attention becomes dispersed due to similar contours and a lack of color cues. The ISTA features then guide the SNN branch to concentrate on the correct object.

\subsection{Ablation Study}
\label{subsec:result-ablation}

\subsubsection{Single Modality vs. Dual Modality}

\cref{tab:ablation-modal} compares the performance of single- and dual-modality models. ``RGB-ANN" and ``Event-SNN" denote models trained only on RGB and event data via the ANN and SNN branches, respectively. Our hybrid ISTASTrack outperforms single-modality models across all four datasets. In particular, Event-SNN surpasses RGB-ANN on the FE240hz dataset. This dataset features extreme lighting conditions that favor event data. In contrast, on the other three datasets, RGB-ANN significantly outperforms Event-SNN and is nearly comparable to the hybrid model, indicating a stronger reliance on the RGB modality.

\begin{table}[htb]
    \centering
    \caption{Comparison of tracking results of single- and dual-modality models. \textbf{Bold} values represent the best results for the same dataset.}
    \label{tab:ablation-modal}
    \setlength{\tabcolsep}{1mm}
    \renewcommand{\arraystretch}{1.2}
\begin{tabular}{l|cc|cc|cc|cc}
\hline
\multirow{2}[4]{*}{\textbf{Model}} & \multicolumn{2}{c|}{\textbf{FE240}} & \multicolumn{2}{c|}{\textbf{VisEvent}} & \multicolumn{2}{c|}{\textbf{COESOT}} & \multicolumn{2}{c}{\textbf{FELT}} \\
\cline{2-9}      & \textbf{SR} & \textbf{PR} & \textbf{SR} & \textbf{PR} & \textbf{SR} & \textbf{PR} & \textbf{SR} & \textbf{PR} \\
\hline
\textbf{RGB-ANN} & 50.9  & 73.5  & 59.3  & 74.2  & 70.2  & 79.8  & 48.1  & 55.9  \\
\textbf{Event-SNN} & 53.8  & 80.7  & 35.5  & 47.4  & 50.0  & 55.8  & 35.2  & 40.2  \\
\textbf{Hybrid} & \textbf{64.6} & \textbf{92.8} & \textbf{61.4} & \textbf{76.8} & \textbf{71.4} & \textbf{81.7} & \textbf{50.1} & \textbf{58.8} \\
\hline
\end{tabular}%
\end{table}

\subsubsection{Backbone Design and ISTA Contribution}

To clarify the source of performance gains, we conduct ablation experiments comparing dual-ANN and hybrid ANN-SNN backbones with varying depths and ISTA adapter usage (\cref{tab:ablation-SNN-D}). The results show that the hybrid design achieves accuracy comparable to dual-ANN models while offering improved energy efficiency (see further analysis in \cref{subsec:result-efficiency}).

We further examine whether architectural asymmetry affects performance by reducing four layers in the ANN branch (D8-D8) and adding four layers to the SNN branch (D12-D12). The D8-D8 configuration causes a clear drop in tracking precision. This highlights the importance of maintaining the pretrained ANN depth. In contrast, D12-D12 yields only marginal improvement, indicating that simply increasing SNN depth does not lead to performance gains.

Overall, these results suggest that the observed improvements mainly arise from the ISTA adapter rather than differences in backbone depth or architecture.

\begin{table}[tb]
     \centering
     \caption{Comparison of tracking performance for dual-ANN and ANN-SNN backbones across varying depths and ISTA adapter configurations. D8-D8 and D12-D12 denote 8 or 12 layers for ANN and SNN backbones, respectively. \textbf{Bold} values represent the best results for the same dataset.}
     \label{tab:ablation-SNN-D}
     % \scriptsize
     \setlength{\tabcolsep}{0.8mm}
     \renewcommand{\arraystretch}{1.3}
\begin{tabular}{c|l|cc|cc|cc|cc}
\hline
      & \multirow{2}[1]{*}{\textbf{Config}} & \multicolumn{2}{c|}{\textbf{FE240}} & \multicolumn{2}{c|}{\textbf{VisEvent}} & \multicolumn{2}{c|}{\textbf{COESOT}} & \multicolumn{2}{c}{\textbf{FELT}} \\
\cline{3-10}      &       & \textbf{SR} & \textbf{PR} & \textbf{SR} & \textbf{PR} & \textbf{SR} & \textbf{PR} & \textbf{SR} & \textbf{PR} \\
\hline
\multirow{2}[2]{*}{\textbf{ANN-ANN}} & \textbf{w/o ISTA} & 61.5  & 90.5  & 58.8  & 74.5  & 71.0  & 81.6  & 49.6  & 58.0  \\
      & \textbf{w/ ISTA} & 64.0  & 92.6  & 60.7  & 75.5  & 71.3  & \textbf{81.7} & \textbf{50.3} & \textbf{59.0} \\
\hline
\multirow{5}[6]{*}{\textbf{ANN-SNN}} & \textbf{w/o ISTA (T=1)} & 60.1  & 88.0  & 59.4  & 74.7  & 70.5  & 80.4  & 48.9  & 57.3  \\
      & \textbf{w/o ISTA (T=3)} & 61.3  & 89.2  & 60.3  & 75.8  & 70.8  & 80.8  & 49.1  & 57.2  \\
\cline{2-10}      & \textbf{D8-D8} & 58.7  & 86.5  & 57.5  & 74.0  & 68.6  & 79.4  & 46.9  & 55.2  \\
      & \textbf{D12-D12} & 64.0  & 93.5  & 60.2  & 75.9  & 71.3  & \textbf{81.7} & 49.7  & 58.3  \\
\cline{2-10}      & \textbf{w/ ISTA, D12-D8} & \textbf{64.6} & \textbf{92.8} & \textbf{61.4} & \textbf{76.8} & \textbf{71.4} & \textbf{81.7} & 50.1  & 58.8  \\
\hline
\end{tabular}%
% Table generated by Excel2LaTeX from sheet 'show-as-others'
 \end{table}

\subsubsection{Fusion Strategy and ISTA Configuration}
 
We conduct an ablation study comparing the proposed ISTA adapters with common feature-level fusion strategies. In addition, we evaluate different ISTA configurations, including the TDA module.

First, we examine several standard fusion approaches, including feature addition, concatenation, and cross-attention. As shown in \cref{tab:ablation-ISTA}, ISTASTrack consistently achieves the best performance across datasets. It shows notable improvements on FE240 and FELT. Interestingly, concatenation and cross-attention do not consistently outperform simple feature addition. This suggests that increased fusion complexity does not necessarily yield better performance.

Compared with the baseline using feature addition, introducing unidirectional adapters (RGB-to-Event or Event-to-RGB) yields clear performance gains, with the Event-to-RGB direction providing slightly larger gains. Employing bidirectional adapters achieve the highest SR and PR scores by enabling mutual feature adaptation between modalities. In addition, removing the TDA module (w/o TDA) leads to reduced accuracy. This result highlights its role in aggregating semantic temporal information from multi-step event features.

  \begin{table}[tb]
     \centering
     % \scriptsize
     \label{tab:ablation-ISTA}
     \caption{Comparison of tracking performance across fusion strategies and ISTASTrack configurations. \textbf{Bold} values represent the best results for the same dataset.}
     \setlength{\tabcolsep}{1mm}
     \renewcommand{\arraystretch}{1.3}
% Table generated by Excel2LaTeX from sheet 'show-as-others'
\begin{tabular}{l|cc|cc|cc|cc}
\hline
\multicolumn{1}{c|}{\multirow{2}[1]{*}{\textbf{Model}}} & \multicolumn{2}{c|}{\textbf{FE240}} & \multicolumn{2}{c|}{\textbf{VisEvent}} & \multicolumn{2}{c|}{\textbf{COESOT}} & \multicolumn{2}{c}{\textbf{FELT}} \\
\cline{2-9}      & \textbf{SR} & \textbf{PR} & \textbf{SR} & \textbf{PR} & \textbf{SR} & \textbf{PR} & \textbf{SR} & \textbf{PR} \\
\hline
\textbf{Add} & 60.1  & 88.0  & 59.4  & 74.7  & 70.5  & 80.4  & 48.9  & 57.3  \\
\textbf{Concat} & 59.8  & 87.0  & 60.1  & 75.8  & 70.4  & 80.4  & 49.2  & 57.5  \\
\textbf{CrossAttn} & 59.2  & 86.5  & 60.5  & 76.1  & 70.8  & 80.9  & 49.1  & 57.5  \\
\hline
\textbf{RGB-to-Event} & 61.7  & 90.5  & 60.4  & 75.9  & 71.2  & \textbf{81.7} & 49.3  & 57.8  \\
\textbf{Event-to-RGB} & 64.3  & \textbf{93.2} & 60.5  & 76.1  & 71.1  & 81.4  & 49.3  & 57.7  \\
\textbf{w/o TDA} & 63.5  & 91.5  & 60.6  & 75.9  & 71.0  & 81.3  & 50.0  & 58.9  \\
\hline
\textbf{Bidirectional} & \textbf{64.6} & 92.8  & \textbf{60.8} & \textbf{76.3} & \textbf{71.4} & \textbf{81.7} & \textbf{50.1} & \textbf{58.8} \\
\hline
\end{tabular}%
 \end{table}

\subsubsection{Adapter Number and Location}

\begin{table}[tb]
    \centering
    \caption{Comparison of tracking performance across different adapter numbers and layers ($T=3$). \textbf{Bold} values represent the best results for the same dataset.}
    \label{tab:ablation-ISTA-num}
    \setlength{\tabcolsep}{1mm}
    % \footnotesize
    \renewcommand{\arraystretch}{1.3}
% Table generated by Excel2LaTeX from sheet 'show-as-others'
\begin{tabular}{l|cc|cc|cc|cc}
\hline
\multicolumn{1}{p{4.125em}|}{\textbf{Model}} & \multicolumn{2}{c|}{\textbf{FE240}} & \multicolumn{2}{c|}{\textbf{VisEvent}} & \multicolumn{2}{c|}{\textbf{COESOT}} & \multicolumn{2}{c}{\textbf{FELT}} \\
\cline{2-9}\textbf{(Num, layer)} & \textbf{SR} & \textbf{PR} & \textbf{SR} & \textbf{PR} & \textbf{SR} & \textbf{PR} & \textbf{SR} & \textbf{PR} \\
\hline
\multicolumn{1}{p{4.125em}|}{\textbf{L=2, 1-2}} & 63.0  & 90.5  & 60.1  & 75.8  & 71.1  & 81.4  & 49.8  & 58.5  \\
\textbf{L=4, 1-4} & \textbf{64.6} & \textbf{92.8} & 60.8  & 76.3  & \textbf{71.3} & \textbf{81.6} & \textbf{50.1} & \textbf{58.8} \\
\textbf{L=6, 1-6} & 63.5  & 91.1  & 59.6  & 74.9  & 70.7  & 81.2  & 49.8  & 58.4  \\
\textbf{L=8, 1-8} & 63.9  & 92.6  & 60.1  & 75.7  & 70.5  & 81.0  & 49.8  & 58.4  \\
\hline
\textbf{L=4, 8-12} & 63.3  & 91.7  & 59.9  & 75.1  & 70.9  & 81.1  & 49.4  & 58.0  \\
\textbf{L=8, 4-12} & 63.0  & 91.2  & \textbf{61.1} & \textbf{76.7} & 70.8  & 80.7  & 49.7  & 58.6  \\
\hline
\end{tabular}%

\end{table}

Since ANN and SNN branches have different transformer encoder depths (12 for ANN and 8 for SNN), both the number and placement of ISTA adapters influence the fusion performance. As shown in \cref{tab:ablation-ISTA-num}, inserting adapters in the first $L$ layers (the first four rows) generally yields higher SR and PR scores than placing them in the last $L$ layers (the last two rows). This suggests that early feature interaction is more effective than late fusion. 

In terms of quantity, adding more adapters does not necessarily yield better results, where four layers of adapters achieve the best overall performance. Once early cross-modal information is exchanged, each branch can effectively process the integrated features independently, whereas additional adapters in deeper layers may interfere with these refined representations. Compared to the model without adapters (see \cref{tab:ablation-ISTA}), incorporating only two layers of ISTA adapters already brings substantial improvements. This demonstrates that effective cross-modal feature adaptation and fusion between ANN and SNN branches.

\subsubsection{Number of Time Steps}

Increasing the number of event time steps $T$ generally preserves richer temporal cues and produces more informative temporal features. We evaluate this effect by varying $T$ for both single-modality Event-SNN and the hybrid ANN-SNN model in \cref{tab:ablation-time-step}, where ``Event-SNN” uses only event input and ``Hybrid” uses RGB-Event input.

\begin{table}[tb]
    \centering
    \caption{Comparison of tracking performance for varying the number of time steps. ``Event-SNN” refers to SNN models that use event data as inputs, whereas ``Hybrid” denotes hybrid networks that take RGB-Event inputs. \textbf{Bold} values represent the best results for the same dataset.}
    \label{tab:ablation-time-step}
    \setlength{\tabcolsep}{1.5mm}
    % \scriptsize
    \renewcommand{\arraystretch}{1.2}
% Table generated by Excel2LaTeX from sheet 'show-as-others'
\begin{tabular}{c|c|cccc}
\Xhline{2\arrayrulewidth}
\multirow{2}[4]{*}{} & \multirow{2}[1]{*}{\textbf{Time step}} & \multicolumn{2}{c|}{\textbf{SNN}} & \multicolumn{2}{c}{\textbf{Hybrid}} \\
\cline{3-6}      &       & \textbf{SR} & \multicolumn{1}{c|}{\textbf{PR}} & \textbf{SR} & \textbf{PR} \\
\hline
\multirow{2}[2]{*}{\textbf{VisEvent}} & \textbf{T=1} & 33.3  & \multicolumn{1}{c|}{45.2} & \textbf{61.4} & \textbf{76.8} \\
      & \textbf{T=3} & \textbf{35.5} & \multicolumn{1}{c|}{\textbf{47.4}} & 60.8  & 76.3  \\
\hline
\multirow{2}[2]{*}{\textbf{COESOT}} & \textbf{T=1} & 44.9  & \multicolumn{1}{c|}{47.2} & 71.3  & 81.6  \\
      & \textbf{T=3} & \textbf{50.0} & \multicolumn{1}{c|}{\textbf{55.8}} & \textbf{71.4} & \textbf{81.7} \\
\hline
\multirow{2}[2]{*}{\textbf{FELT}} & \textbf{T=1} & 30.0  & \multicolumn{1}{c|}{31.4} & \textbf{50.1} & 58.7  \\
      & \textbf{T=3} & \textbf{35.2} & \multicolumn{1}{c|}{\textbf{40.2}} & \textbf{50.1} & \textbf{58.8} \\
% \Xhline{2\arrayrulewidth}
\bolddoubleline
\multirow{5}[4]{*}{\textbf{FE240 (SNN)}} &       & \textbf{SR} & \textbf{OP50} & \textbf{OP75} & \textbf{PR} \\
\cline{2-6}      & \textbf{T=1} & 52.0  & 65.2  & 23.7  & 79.8  \\
      & \textbf{T=3} & 53.8  & 68.6  & \textbf{27.8} & 80.7  \\
      & \textbf{T=5} & 55.4  & 69.0  & 27.5  & 85.8  \\
      & \textbf{T=7} & \textbf{56.3} & \textbf{69.2} & 27.6  & \textbf{88.0} \\
\hline
\multirow{4}[2]{*}{\textbf{FE240 (Hybrid)}} & \textbf{T=1} & 63.8  & 81.7  & 39.8  & 91.8  \\
      & \textbf{T=3} & 64.6  & 82.3  & 41.6  & \textbf{92.8} \\
      & \textbf{T=5} & \textbf{64.7} & \textbf{82.3} & \textbf{43.3} & 92.2  \\
      & \textbf{T=7} & 63.9  & 81.7  & 41.6  & 91.6  \\
\Xhline{2\arrayrulewidth}
\end{tabular}%
\end{table}

Across the four datasets, increasing $T$ from 1 to 3 significantly improves SR and PR for Event-SNN. However, this trend is not consistently reflected in the hybrid model except for FE240hz. The likely reason is that RGB already performs strongly on VisEvent, COESOT, and FELT, while their event signals are relatively weaker, as shown in \cref{tab:ablation-modal}. In contrast, FE240hz provides more balanced modalities, where event inputs can outperform RGB alone. Consequently, improvements in Event-SNN translate into hybrid gains primarily on FE240hz.

We further test larger $T$ values ($T=5,7$) on FE240hz. Increasing $T$ improves both Event-SNN and hybrid performance. Notably, the OP75 metric increases from 39.8\% to 43.3\% as $T$ grows from 1 to 5. However, the improvement becomes marginal at larger $T$. The hybrid model even degrades slightly at $T=7$. Considering the additional computational and memory cost, $T=3$ provides a favorable trade-off while effectively leveraging multi-step SNN dynamics.

\subsection{Computational Efficiency}
\label{subsec:result-efficiency}

\cref{tab:results-energy} compares computational costs and operation breakdown across our ISTASTrack variants and representative RGB-Event trackers. We evaluate the inference speed on an NVIDIA GeForce RTX 4090 GPU and an Intel Xeon Gold 6326 CPU @ 2.90GHz. The upper part reports model size, FPS, FLOPs, and estimated energy for ANN-only (BAT, TENet), hybrid ANN-SNN (MMHT), and SNN-based (SpikeFET) models. Our hybrid ANN-SNN model has a relatively large parameter count due to separate transformer backbones for ANN and SNN, whereas other dual-branch trackers like BAT and SpikeFET share weights between branches. In contrast, our dual-ANN model with shared weights reduces the parameter count (158.0M vs. 92.5M) and increases inference speed (32.9 vs. 65.5 FPS). This model achieves comparable FLOPs and parameters to BAT but runs faster (65.5 vs. 59.6 FPS). However, models with SNN backbones like our ANN-SNN, MMHT, and SpikeFET show lower speeds of roughly 20-30 FPS. This occurs because current hardware does not fully optimize SNN operations. Despite this limitation, the ANN-SNN model substantially lowers estimated energy consumption. These results demonstrate its potential for energy-efficient deployment.

%Models incorporating SNN backbones, such as our ANN-SNN, MMHT, and SpikeFET, exhibit reduced FPS ($\sim$20-30) primarily because the current implementations of SNN operations are not fully optimized for GPU or CPU execution. Despite this, the ANN-SNN model substantially lowers estimated energy consumption compared with dual-ANN trackers, demonstrating its potential for energy-efficient deployment.

% Direct comparisons across different architectures are not entirely fair, as parameter count is largely determined by backbone design. 

To isolate the effect of our modules, we compare ISTASTrack with and without ISTA adapters under the same architecture, and report the operation breakdown in the lower part of \cref{tab:results-energy}. The ISTA adapter adds only 0.32M parameters ($\sim$2\% of the total parameters). It causes a minor reduction in inference speed (37.5 FPS vs. 32.9 FPS) and marginal additional energy cost. The TDA module is even more lightweight. Its contribution to computational cost is practically negligible. These results show that our modules efficiently enable cross-modal and cross-paradigm fusion. They provide consistent performance gains with minimal overhead. Consequently, our approach achieves a favorable trade-off between accuracy and efficiency.

\begin{table}[tb]
     \centering
     \caption{Comparison of computational costs and operation breakdown. Upper: overall efficiency of different RGB-Event trackers. Lower: operation breakdown of our model.
    ``ANN-SNN” denotes the hybrid ISTASTrack,``ANN-ANN” replaces the SNN branch with ANN. ``FLOPs*" represents the summation of FLOPs and SyOps.} 
     \label{tab:results-energy}
     % \scriptsize
     \setlength{\tabcolsep}{1mm}
\renewcommand{\arraystretch}{1.2}
\begin{tabular}{cc|c|cc|c}
\Xhline{2\arrayrulewidth}
% \toprule
\multicolumn{2}{c|}{\multirow{2}[1]{*}{\textbf{Model}}} & \multirow{2}[1]{*}{\textbf{Params}} & \multicolumn{1}{c|}{\multirow{2}[1]{*}{\textbf{FPS}}} & \multirow{2}[1]{*}{\textbf{FLOPs*(G)}} & \multirow{2}[1]{*}{\textbf{E(mJ)}} \\
\multicolumn{2}{c|}{} &       & \multicolumn{1}{c|}{} &       &  \\
\hline
\multicolumn{2}{c|}{\textbf{TENet}} & 130.3M & \multicolumn{1}{c|}{74.4} & 149.4  & 149.4  \\
\multicolumn{2}{c|}{\textbf{BAT}} & 92.4M & \multicolumn{1}{c|}{59.6} & 112.9  & 259.7  \\
\multicolumn{2}{c|}{\textbf{MMHT}} & 14.0M & \multicolumn{1}{c|}{24.5} & 9.8   & 22.5  \\
\multicolumn{2}{c|}{\textbf{SpikeFET}} & 105.5M & \multicolumn{1}{c|}{26.2} & 53.2  & 104.8  \\
\hline
\multicolumn{2}{c|}{\textbf{ANN-ANN}} & 92.5M & \multicolumn{1}{c|}{65.5} & 112.7  & 258.9  \\
\cline{1-2}
\multicolumn{1}{c|}{\multirow{3}[1]{*}{\textbf{ANN-SNN}}} & \multicolumn{1}{l|}{\textbf{ T=1, w/o ISTA}} & 157.7M & \multicolumn{1}{c|}{37.5} & 67.8  & 144.2  \\
\multicolumn{1}{c|}{} & \multicolumn{1}{l|}{\textbf{ T=1}} & 158.05M & \multicolumn{1}{c|}{32.9} & 66.2  & 146.0  \\
% \cline{3-6}
\multicolumn{1}{c|}{} & \multicolumn{1}{l|}{\textbf{ T=3}} & 158.05M & \multicolumn{1}{c|}{21.2} & 69.8  & 169.1  \\
% \Xhline{2\arrayrulewidth}
\bolddoubleline
% \midrule[1pt]  % 稍细的双线第一条
% \midrule[1pt]  % 稍细的双线第二条
\multicolumn{2}{c|}{\multirow{2}[1]{*}{\textbf{Model}}} & \multirow{2}[1]{*}{\textbf{Params}} & \multicolumn{2}{c|}{\textbf{FLOPs(G)}} & \multirow{2}[1]{*}{\textbf{SyOps(G)}} \\
\cline{4-5}\multicolumn{2}{c|}{} &       & \textbf{ANN} & \textbf{SNN} &  \\
\hline
\multicolumn{2}{c|}{\textbf{ANN-SNN (T=1)}} & 158.05M & 58.3  & 3.7   & 4.2  \\
\hline
\multicolumn{2}{c|}{\textbf{ANN-SNN (T=3)}} & 158.05M & 58.5  & 10.9  & 11.4  \\
\multicolumn{2}{c|}{\textbf{ISTA Adapter}} & 0.32M & 0.31  & -     & - \\
\multicolumn{2}{c|}{\textbf{TDA}} & 81    & 0.0032  & -     & - \\
\Xhline{2\arrayrulewidth}
\end{tabular}%
 \end{table}

\section{Limitation}

First, the SNN branch currently do not fully exploit their potential for modeling long dynamic sequences. This limitation stems from memory constraints. We currently reset SNN states after each inference step. This prevents the network from leveraging long-term temporal dependencies.  Future work will explore mechanisms to maintain and update temporal context across longer sequences for high-speed vision.

Second, the architecture relies on pretrained backbones with different depths for the ANN and SNN branches. This design leads to a relatively large model size and constrains performance to the capacity of the chosen backbones. Future work could extend the framework to lighter or more efficient backbones, such as ConvFormer or Swin-based variants with downsampled windows. These changes may also enhance performance on small objects.

Finally, the current energy efficiency analysis is based on estimated hardware operation costs rather than direct neuromorphic measurements. Although hardware-level tools such as CACTI \cite{li_cactip_2011,liang_h2learn_2022, qi_ebrainisa_2026} can be used to estimate memory and system-level energy overhead, integrating such evaluations requires dedicated system-level implementation and is left for future work.

\section{Conclusion}
\label{sec:conclusion}

In this work, we propose ISTASTrack, a hybrid ANN-SNN tracker with ISTA adapters for RGB-Event VOT. The hybrid network combines a vision transformer for RGB inputs and a spiking transformer for event streams, leveraging the complementary strengths of the two modalities. Between transformer encoders in the ANN and SNN branches, we design ISTA adapters based on sparse representation theory and algorithm unfolding. These lightweight ISTA adapters effectively achieve bidirectional feature interaction and adaptation. Additionally, a temporal downsampling attention (TDA) module ensures semantic temporal alignment between multi-step SNN features and single-step ANN features. Experimental results demonstrate that ISTASTrack achieves high tracking accuracy and improved energy efficiency across multiple datasets and challenging scenarios. Furthermore, our comprehensive analyses provide guidance for for the effective fusion of ANN and SNN paradigms and designing hybrid networks..

\bibliographystyle{IEEEtran} %
\bibliography{references}

\vspace{-20pt}

\begin{IEEEbiography}[{\includegraphics[width=1in,height=1.25in,clip,keepaspectratio]{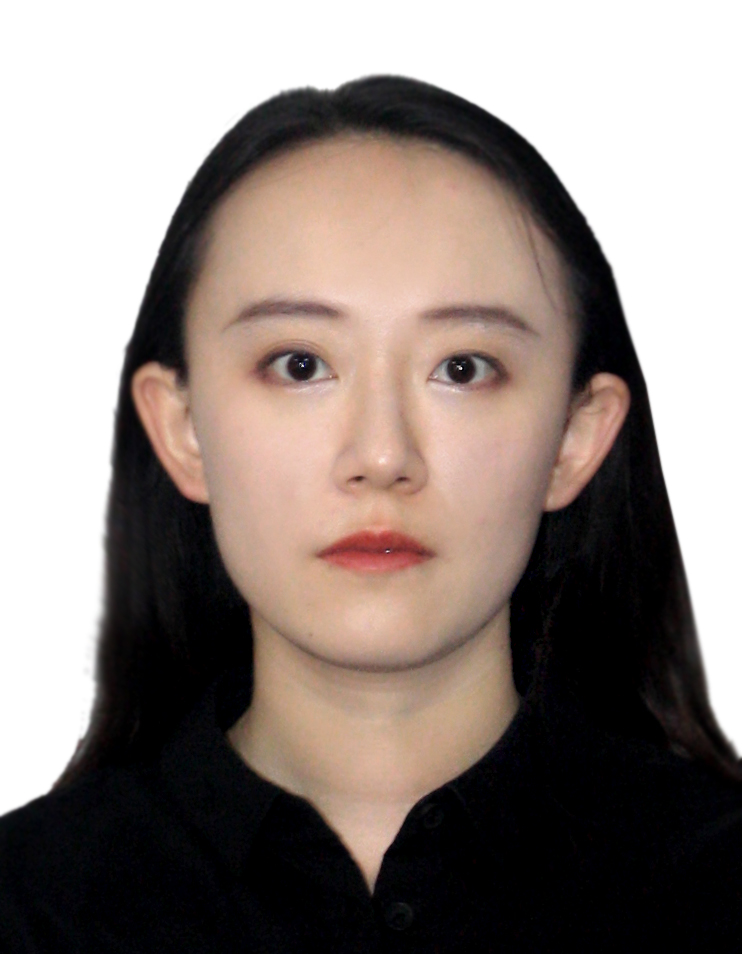}}]{Siying Liu}
received the Bachelor's and Master's degrees in Instrumentation Science and Technology from Beihang University, Beijing, China, in 2017 and 2020, respectively. In 2024, she received her Ph.D. degree at the Department of Electrical and Electronic Engineering, Imperial College London, U.K. She is currently a postdoctoral researcher at the Center for Brain Inspired Computing Research (CBICR), Tsinghua University, Beijing, China, and has been selected for the Shuimu Tsinghua Scholar Program. Her research interests include event-based vision, brain-inspired computing, and machine learning.
\end{IEEEbiography}

\vspace{-20pt}

\begin{IEEEbiography}[{\includegraphics[width=1in,height=1.25in,clip,keepaspectratio]{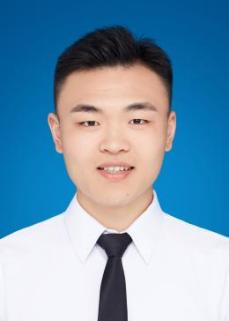}}]{Zikai Wang} received the M.S. degree from Jiangxi University of Science and Technology, Ganzhou, China, in 2023. He is currently pursuing the Ph.D. degree with the College of Computer Science and Technology, Taiyuan University of Technology, Taiyuan, China. 
His current research interests include brain-inspired computing models and applications, with a focus on RGB-Event tracking task, multi-modal learning and fusion.
\end{IEEEbiography}

\vspace{-20pt}

\begin{IEEEbiography}[{\includegraphics[width=1in,height=1.25in,clip,keepaspectratio]{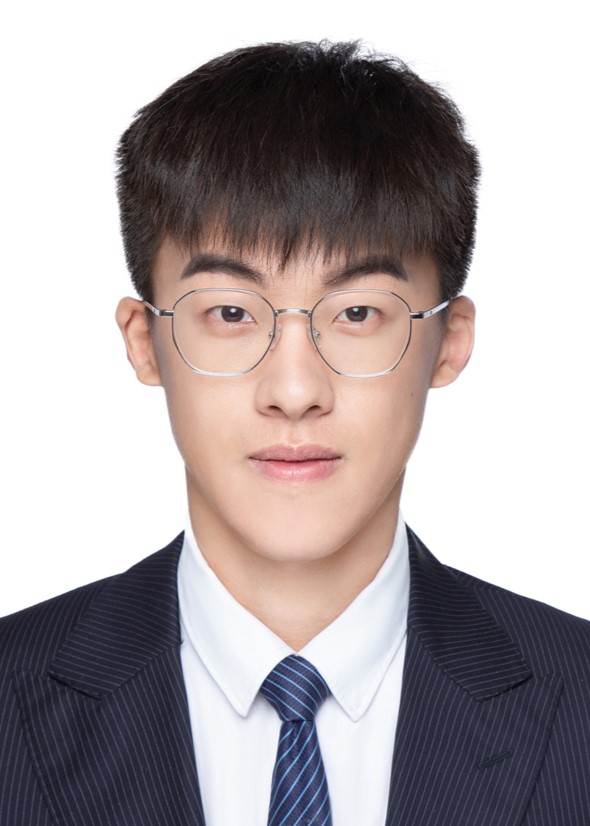}}]{Hanle Zheng} received the B.E. degree in Measurement and Control Technology and Instruments from the Department of Precision Instrument, Tsinghua University, Beijing, China, in 2022. He is currently pursuing the Ph.D. degree with the Department of Precision Instrument, Tsinghua University, Beijing, China. His research interests include brain-inspired learning, spiking neural networks and dynamic system.
\end{IEEEbiography}

\vspace{-20pt}

\begin{IEEEbiography}[{\includegraphics[width=1in,height=1.25in,clip,keepaspectratio]{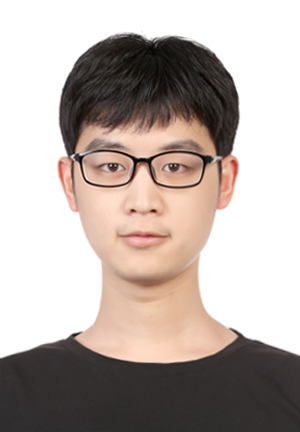}}]{Yifan Hu} received the B.S. degree from Tsinghua University, Beijing, China, in 2019. He received the Ph.D. degree from the Center for Brain Inspired Computing Research (CBICR), Department of Precision Instrument, Tsinghua University, in 2025. His research interests include deep learning and neuromorphic computing.
\end{IEEEbiography}

\vspace{-20pt}

\begin{IEEEbiography}[{\includegraphics[width=1in,height=1.25in,clip,keepaspectratio]{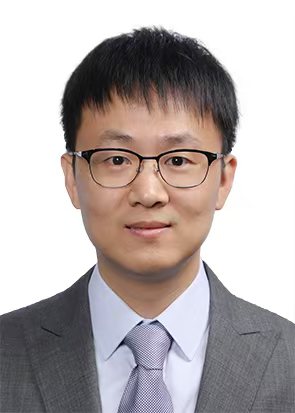}}]{Xilin Wang} received B.S. degree in from Department of Materials Science and Technology, Beijing University of Chemical Technology and Ph.D degree from Department of Materials Science and Technology, Tsinghua University. Now he is an Associate Professor at the Shenzhen Graduate School of Tsinghua University. His research areas include high-voltgae insulation and dielectric materials, and the interdisciplinary fields between materials science, electrical engineering and AI.
\end{IEEEbiography}

\vspace{-10pt}

\begin{IEEEbiography}[{\includegraphics[width=1in,height=1.25in,clip,keepaspectratio]{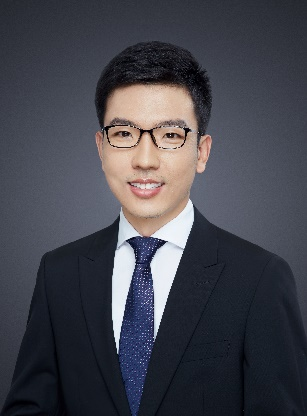}}]{Qingkai Yang} (Senior Member, IEEE) received the first Ph.D. degree in control science and engineering from Beijing Institute of Technology, Beijing, China, in 2018, and the second Ph.D. degree in system control from the University of Groningen, Groningen, The Netherlands, in 2018. He is currently a Professor with the School of Automation, Beijing Institute of Technology. His research interests include the morphological intelligence, autonomous mobile robots, and cooperative control of multi-agent systems.
\end{IEEEbiography}

\vspace{-10pt}

\begin{IEEEbiography}[{\includegraphics[width=1in,height=1.25in]{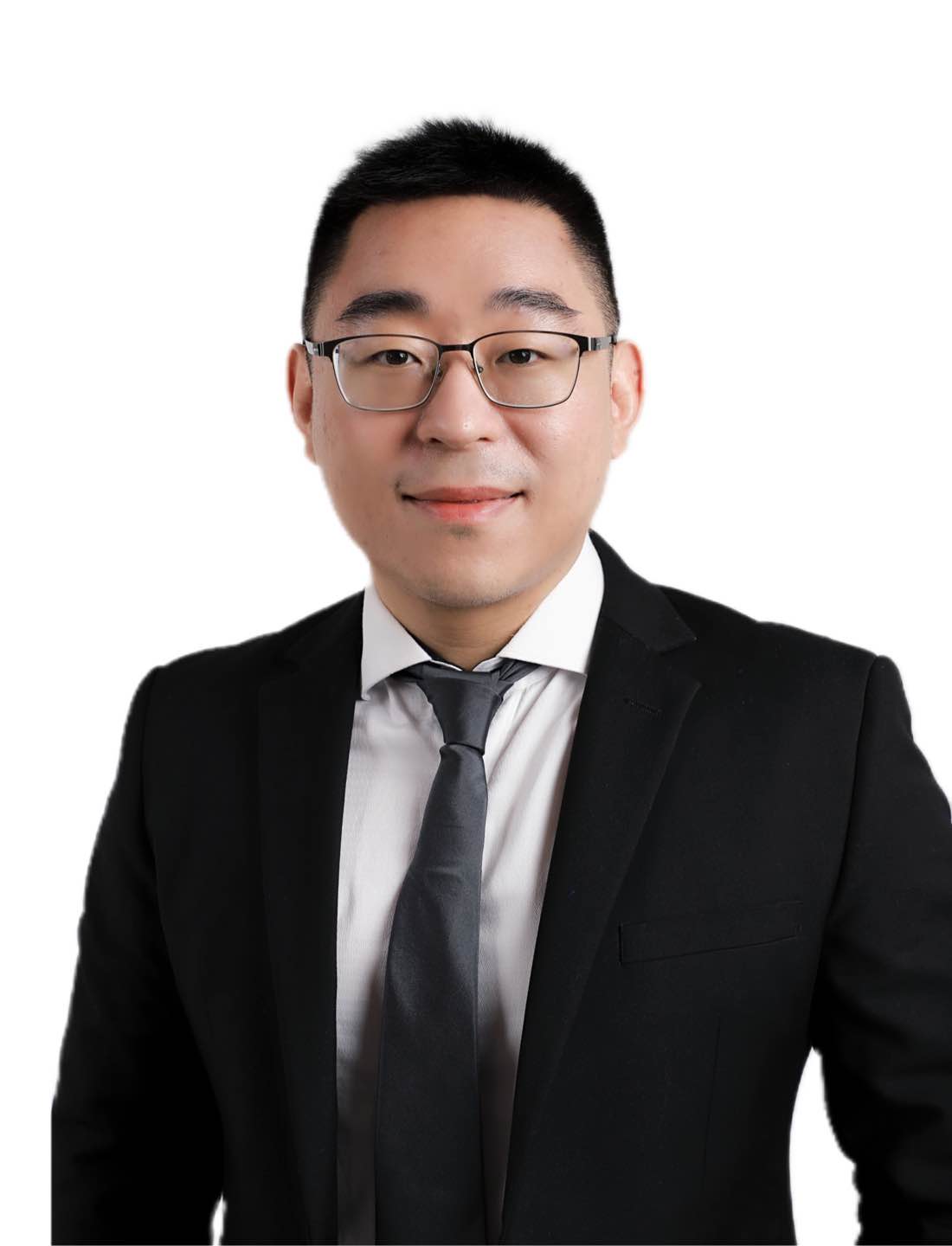}}]{Jibin Wu} (Member, IEEE)
received the B.E. and Ph.D degree in Electrical Engineering from National University of Singapore, Singapore in 2016 and 2020, respectively. Dr. Wu is currently an Assistant Professor in the Department of Data Science and Artificial Intelligence, The Hong Kong Polytechnic University. His research interests broadly include brain-inspired artificial intelligence, neuromorphic computing, computational audition, speech processing, and machine learning. He is currently serving as the Associate Editors for IEEE Transactions on Neural Networks and Learning Systems and IEEE Transactions on Cognitive and Developmental Systems.
\end{IEEEbiography}

%Dr. Wu has published over 40 papers in prestigious conferences and journals in artificial intelligence and speech processing, including NeurIPS, ICLR, AAAI, TPAMI, TNNLS, TASLP, and IEEE JSTSP.
\vspace{-10pt}

\begin{IEEEbiography}[{\includegraphics[width=1in,height=1.25in,clip,keepaspectratio]{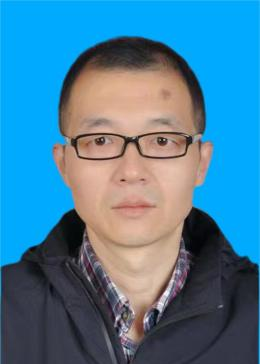}}]{Hao Guo} received the Ph.D. degree from the College of Computer Science and Tecnology, Taiyuan University of Technology, Taiyuan, China, in 2013. 
He is currently a Full Professor with Taiyuan University of Technology, Taiyuan, China. His current research interests include brain-inspired computing and artificial intelligence. He is an ACM member and a Senior Member of the China Computer Federation.
\end{IEEEbiography}

\vspace{-20pt}

\begin{IEEEbiography}[{\includegraphics[width=1in,height=1.25in,clip,keepaspectratio]{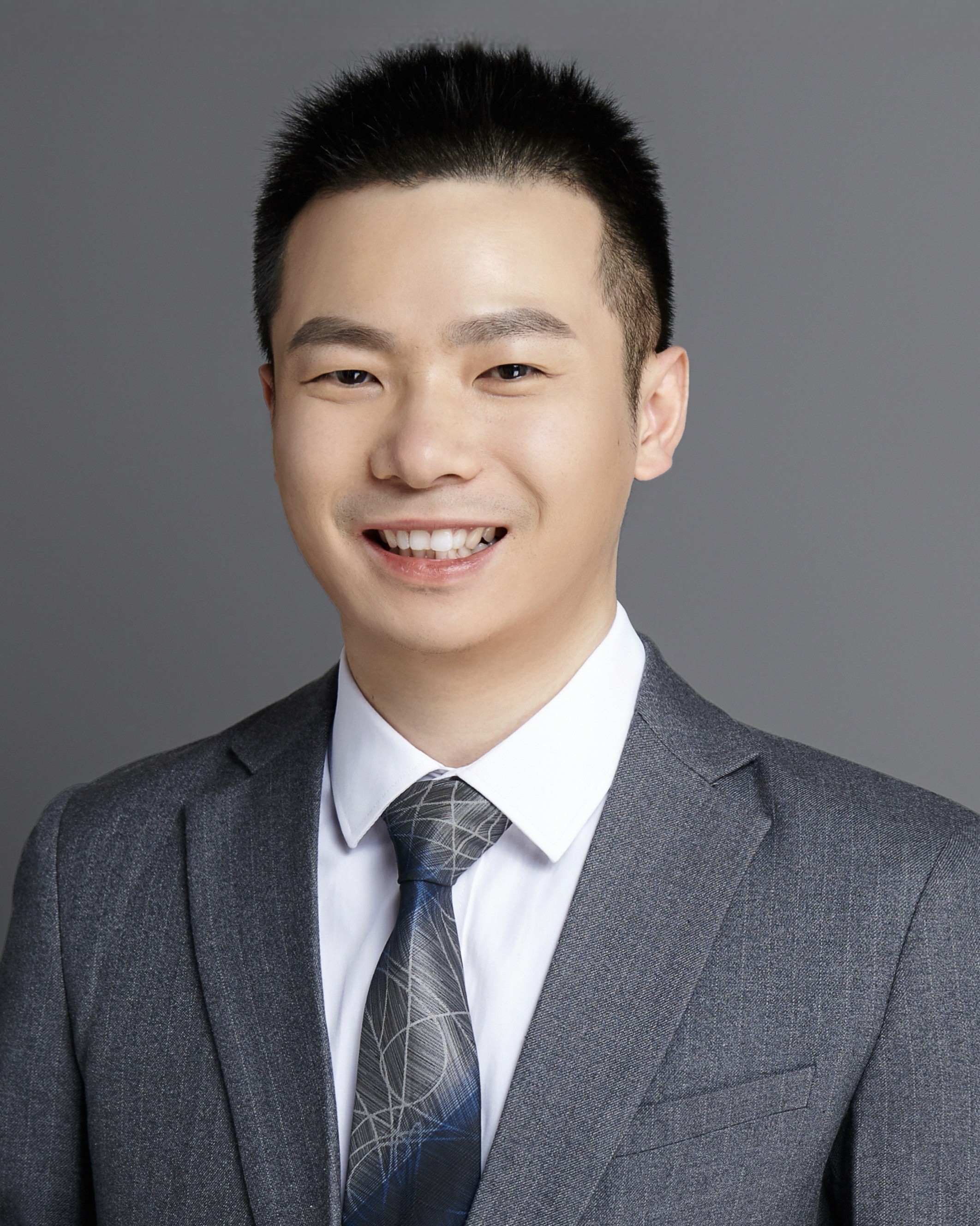}}]{Lei Deng} (Senior Member, IEEE) received the B.E. degree from University of Science and Technology of China in 2012 and the Ph.D. degree from Tsinghua University in 2017. He was a Postdoctoral Fellow at University of California, Santa Barbara from 2017 to 2021. He is currently an Associate Professor at Tsinghua University, Beijing, China. He has focused on studying brain-inspired models, chips, and applications for more than 13 years. He has published over 100 papers in prestigious journals/conferences such as Nature, Nature Communications, Proc. IEEE, IEEE TPAMI, and IEEE JSSC, including over 10
Cover Stories, Highly-Cited Papers, Popular Papers, Featured Articles, and Best Papers. He was a recipient of the Young Scholar of Chinese Institute for Brain Research, Beijing, Outstanding Youth Award of Chinese Association for AI, MIT TR 35 China, and Intel China Academic Achievement Award
(Outstanding Research). He serves as an Associate Editor for Frontiers in Neuroscience and Edge Intelligence and Systems, and a Session Chair or a PC Member for a number of conferences.
\end{IEEEbiography}



\vfill

\end{document}